\documentclass[11pt,letterpaper]{article}
\usepackage{arxiv}

\usepackage[pdftex]{graphicx}
\usepackage{graphicx}
\usepackage{rotate}
\usepackage{wrapfig}
\usepackage{caption}
\usepackage[T1]{fontenc}
\usepackage[english]{babel}
\usepackage{times}
\usepackage{amsmath,amstext,amsbsy,amsopn,amsthm,amsfonts,amssymb}
\usepackage{mathtools}
\usepackage{url}
\usepackage{tabularx,booktabs}
\newcolumntype{C}{>{\centering\arraybackslash}X}
\usepackage{booktabs,array,colortbl}
\usepackage{lastpage}
\usepackage{multirow}
\usepackage{algorithm}
\usepackage{balance}
\usepackage{placeins}
\usepackage{array}
\usepackage{blindtext}
\usepackage{subfigure}
\usepackage{caption}
\usepackage{capt-of}
\usepackage{booktabs}% for better rules in the table
\usepackage[small]{titlesec}
\usepackage{paralist}
\usepackage{verbatim} % For comment
\usepackage[usenames,dvipsnames,svgnames]{xcolor}
\usepackage{algorithmic}
\usepackage{fancyhdr}
\usepackage{wrapfig}
\begin{document}

\title{iPal: A Machine Learning based Smart Healthcare Framework for Automatic Diagnosis of Attention Deficit/Hyperactivity Disorder (ADHD)}

\author{
	\begin{tabular}{ccc}
		Abhishek~Sharma & Arpit~Jain & Shubhangi~Sharma\\
		Dept. of Electronics and&Dept. of Computer&Dept. of Electronics and\\
		Communication Engineering,&Science and Engineering,& Communication Engineering,\\
		The LNMIIT Jaipur, India&The LNMIIT Jaipur, India&The LNMIIT Jaipur, India\\
		\vspace{0.8cm}
		abhisheksharma@lnmiit.ac.in &18ucs085@lnmiit.ac.in&18uec006@lnmiit.ac.in\\
		Ashutosh~Gupta&Prateek~Jain &Saraju P. Mohanty\\
		Dept. of Computer&Dept. of Electronics and&Dept. of Computer\\
		Science and Engineering,& Instrumentation Engineering,& Science and Engineering,\\
		 The LNMIIT Jaipur, India&Nirma University Ahmedabad, India&University of North Texas, USA\\
		18ucs200@lnmiit.ac.in&prtk.ieju@gmail.com&saraju.mohanty@unt.edu\\
	\end{tabular}
}
\maketitle

\cfoot{Page -- \thepage-of-\pageref{LastPage}}

\begin{abstract}
ADHD is a prevalent disorder among the younger population. Standard evaluation techniques currently use evaluation forms, interviews with the patient, and more. However, its symptoms are similar to those of many other disorders like depression, conduct disorder, and oppositional defiant disorder, and these current diagnosis techniques are not very effective. Thus, a sophisticated computing model holds the potential to provide a promising diagnosis solution to this problem. This work attempts to explore methods to diagnose ADHD using combinations of multiple established machine learning techniques like neural networks and SVM models on the ADHD200 dataset and explore the field of neuroscience. In this work, multiclass classification is performed on phenotypic data using an SVM model. The better results have been analyzed on the phenotypic data compared to other supervised learning techniques like Logistic regression, KNN, AdaBoost, etc. In addition, neural networks have been implemented on functional connectivity from the MRI data of a sample of 40 subjects provided to achieve high accuracy without prior knowledge of neuroscience. It is combined with the phenotypic classifier using the ensemble technique to get a binary classifier. It is further trained and tested on 400 out of 824 subjects from the ADHD200 data set and achieved an accuracy of 92.5\% for binary classification The training and testing accuracy has been achieved upto 99\% using ensemble classifier. 

\end{abstract}
%
%%\begin{Keywords}
%%\textbf{Index Terms} - ADHD, Neuro-Imaging, Artificial Neural Network, Ensembling
%%\end{Keywords}
%
%
%
%\IEEEpeerreviewmaketitle
%
\section{Introduction}

\begin{figure*}[htbp]
	\centering
	\includegraphics[width=0.6\textwidth]{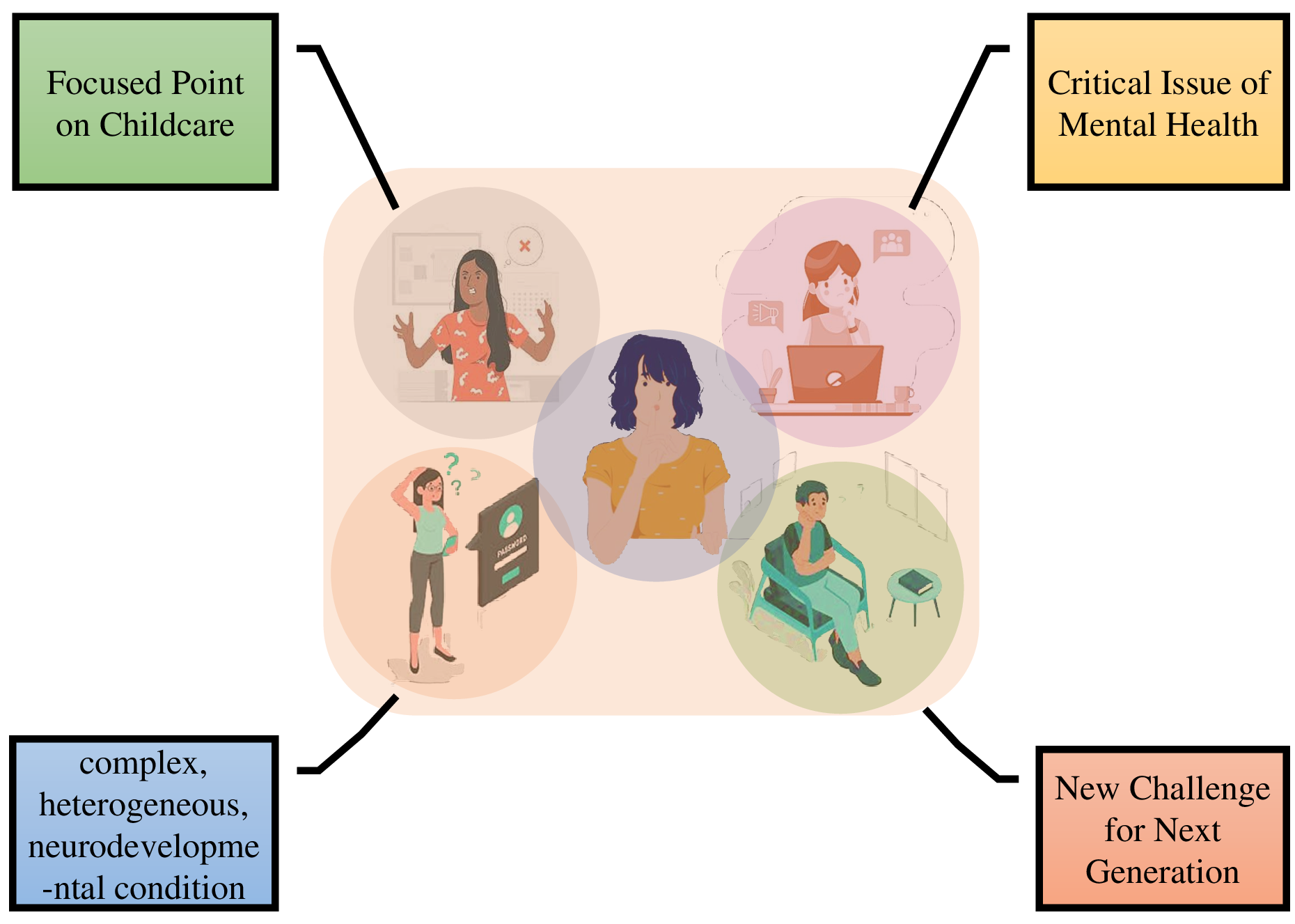}
	\DeclareGraphicsExtensions.
	\caption{An ADHD Symptoms with future Challenges }
	\label{thematic1}
\end{figure*}

The Attention Deficit/Hyperactivity Disorder (ADHD) is a pervasive neurodevelopmental disorder affecting the younger population. However, ADHD does also affects numerous adults [1]. People diagnosed with ADHD generally experience inattention (unable to focus properly), hyperactivity (excessive movement that is not fit for the sitting), and impulsive behaviors (may act without thinking about what the result will be). Mental disorder symptoms include difficulty in focusing, instant irritation, easily distraction and other abnormal mental situations [2, 3]. It is critical issue of mental health and it is being challenged for the next generation. Now a days, it is required to cure the mental disorder efficiently without any constraints of awareness, time and availability of medical experts. Remote connectivity of medical expert and patients is the trendy solution in terms of better and advanced facilities to the patients. The current issues with symptoms are presented in Figure \ref{thematic1}.\\
As the current healthcare approaches are being updated, there is a trend of consumer awareness for their health. In particular scenario, the demand for remote healthcare is getting promoted than ever. Present IoMT framework for healthcare encourages health centers to ameliorate the quality treatment with focusing on overall optimization in terms of the cost also [4]. The ADHD diagnosis is required to be updated as the living style is being sophisticated and patients need the treatment instantly without any basic hurdles to get the treatment. Hence, an ADHD care framework is required to proposed for smart healthcare, in which patient will get support from medical representatives without wastage of time and treatment will be reliable as the patient can approach to the experts remotely and time won't be constraint [5].

\section{Related Works}
According to survey analysis, overall 2.2\% of the average prevalence of ADHD has been estimated in children and adolescents (aged <18 years) [6]. Many studies have been conducted based on machine learning to diagnose ADHD effectively. To contribute to the research in diagnosing ADHD, the ADHD-200 consortium globally held the ADHD-200 competition backed by the International Neuroimaging Data-sharing Initiative (INDI). Evolution of this research is considered, when statistical analysis of brain surfaces has been done using gaussian random field [7]. Adjacently, fMRI images have been taken as resource for ADHD detection in terms of technology advancements. In continuation, EEG signals were being monitored for gesture recognition to justify the changes in brain signals. The research work had been done to optimize the signal parameters to make precise framework. Different features had been extracted for chasing the accuracy for ADHD level justifications. But, it is required to form a reliable framework, which confirm the levels of mental disorders.\\
The prior presented work uses SVM, explicitly addressing the imbalanced dataset problem of ADHD-200 [8]. The positive and negative empirical errors are handled explicitly and separately by using a three-objective SVM. After trying many traditional classifiers and comparing them with deep learning-based CNN VGG 16, the maximum accuracy was achieved in CNN VGG 16 model [9]. 
The earlier presented work demonstrates the use of functional-Magnetic Resonance Imaging for the diagnosis of ADHD using multiple machine learning models on the publicly available ADHD-200 dataset [10-12]. The results show the classification of ADHD and control subjects, differentiate between the functional connectivity of these two categories, and evaluate the significance of phenotypic data to predict ADHD [13]. They have used SVM to classify ADHD after calculating functional connectivity, performing elastic net-based feature selection and integrating phenotypic information. In the presented work, multiclass classification has been performed using a hierarchical extreme learning machine (H-ELM) classifier [14]. They have also compared the performance of the H-ELM classifier with that of a support vector machine and primary extreme learning machine (ELM) classifiers for cortical MRI data from 159 ADHD patients of the ADHD-200 dataset. This work achieved an accuracy of ~61\% by using SVM with recursive feature elimination(RFE-SVM). Overall, they achieved high multiclass classification accuracy by combining RFE-SVM with H-ELM classifiers for s-MRI data. After doing an overview of many studies on ADHD prediction, also found that SVM and ANN are most effective classification techniques [15].

\begin{figure*}[htbp]
	\centering
	\includegraphics[width=0.9\textwidth]{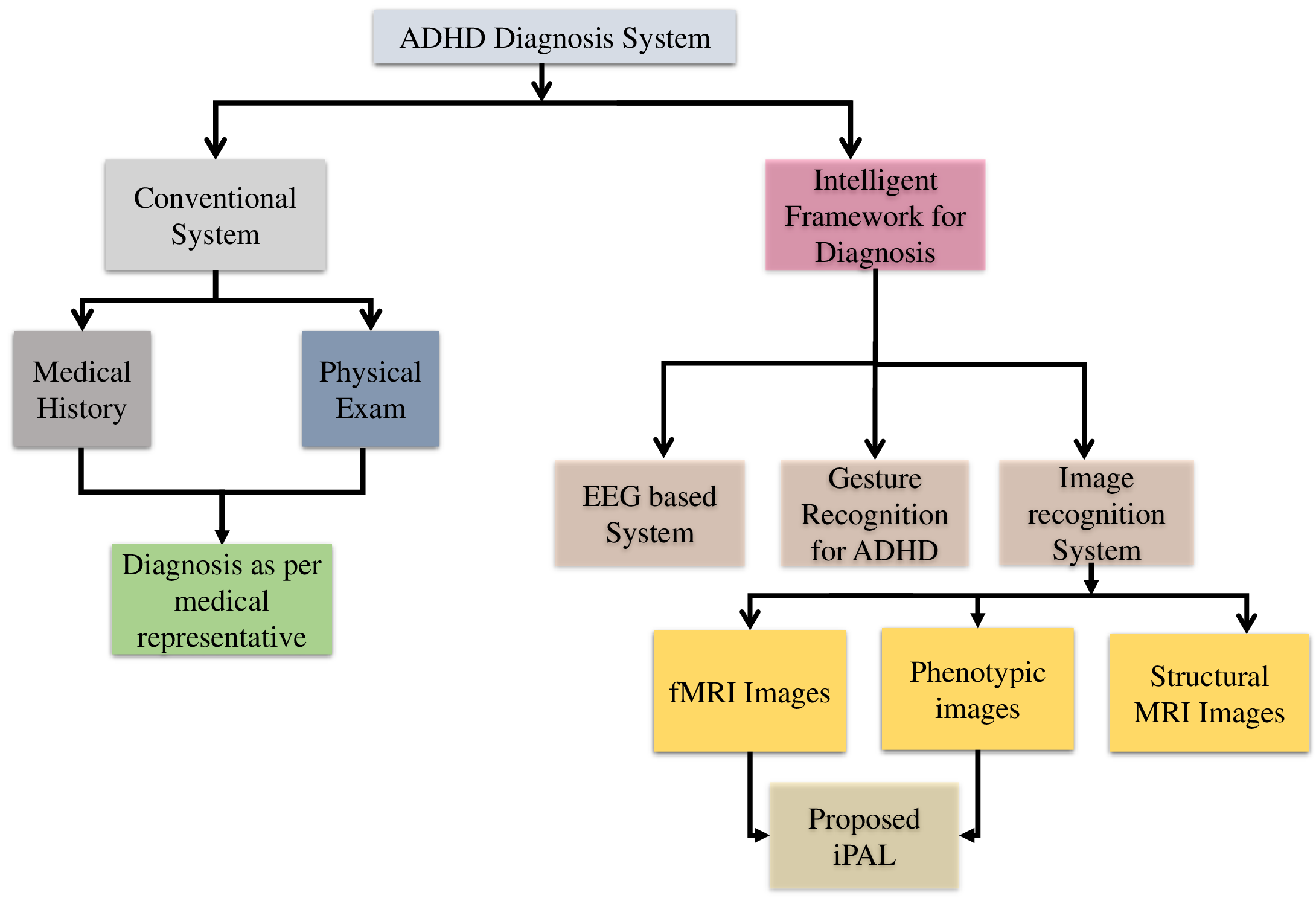}
	\DeclareGraphicsExtensions.
	\caption{Related Prior Works for ADHD}
	\label{prior}
\end{figure*}

The work proposes a Multichannel Deep Neural Network Model, which has delivered a promising result with an accuracy of ~95\% on the combined data of connectome and phenotypic data of the ADHD-200 dataset [16]. Furthermore, the multichannel deep neural network model improved ADHD detection performance considerably compared with a single scale by fusing the multi scale brain connectome data [16]. The work establishes that s-MRI data can effectively differentiate between ADHD and controls [17]. In this work, deep learning neural networks have been used to determine the similarities in neuroanatomical changes in the brains of children with ADHD and adults with ADHD. Not only this, but this work also demonstrates the effective use of neural networks as classification models to test hypotheses about developmental continuity and to predict ADHD. The work proposes a deep neural network-multilayer perceptron to diagnose psychotic disorder diseases (PDD) [18]. The prior works evident that neural networks can be effective in ADHD prediction [9, 16, 18].
The presented figure depicts the earlier technology to the recent technique to detect the mental disorder. These hierarchical report is presented in Figure \ref{prior}.
Still, accuracy and reliability are the recent challenges for future perspective. The main motivation behind the method developed in this study was to explore the efficient supervised leaning techniques in the field of neuroscience and achieve maximum accuracy while at the same time also combine the results of both kinds of data, i.e., Phenotypic and MRI data.
Despite many such studies, this remains open to further research and a topic of interest to date as many studies are still going on and the ADHD-200 dataset still remains a significant contribution to these studies.

\section{Contributions of the Current Paper}
\subsection{The Problem Addressed in the Current Paper}
Presently, the mental disorder is being cured by the psychiatrists after taking counseling and discussing with them. Based on medical and personal histories, experts would be able to get the level of mental disorders and medical treatments are being provided accordingly. There is a disadvantage of this convention process that the patient may not be able to consult with expert sometimes and treatment wouldn't be proper and reliable. Many time, the experts are not available at particular area. So, the problem would be critical in those cases.

\subsection{Solution Proposed in the Current Paper}
\begin{figure*}[htbp]
	\centering
	\includegraphics[width=0.9\textwidth]{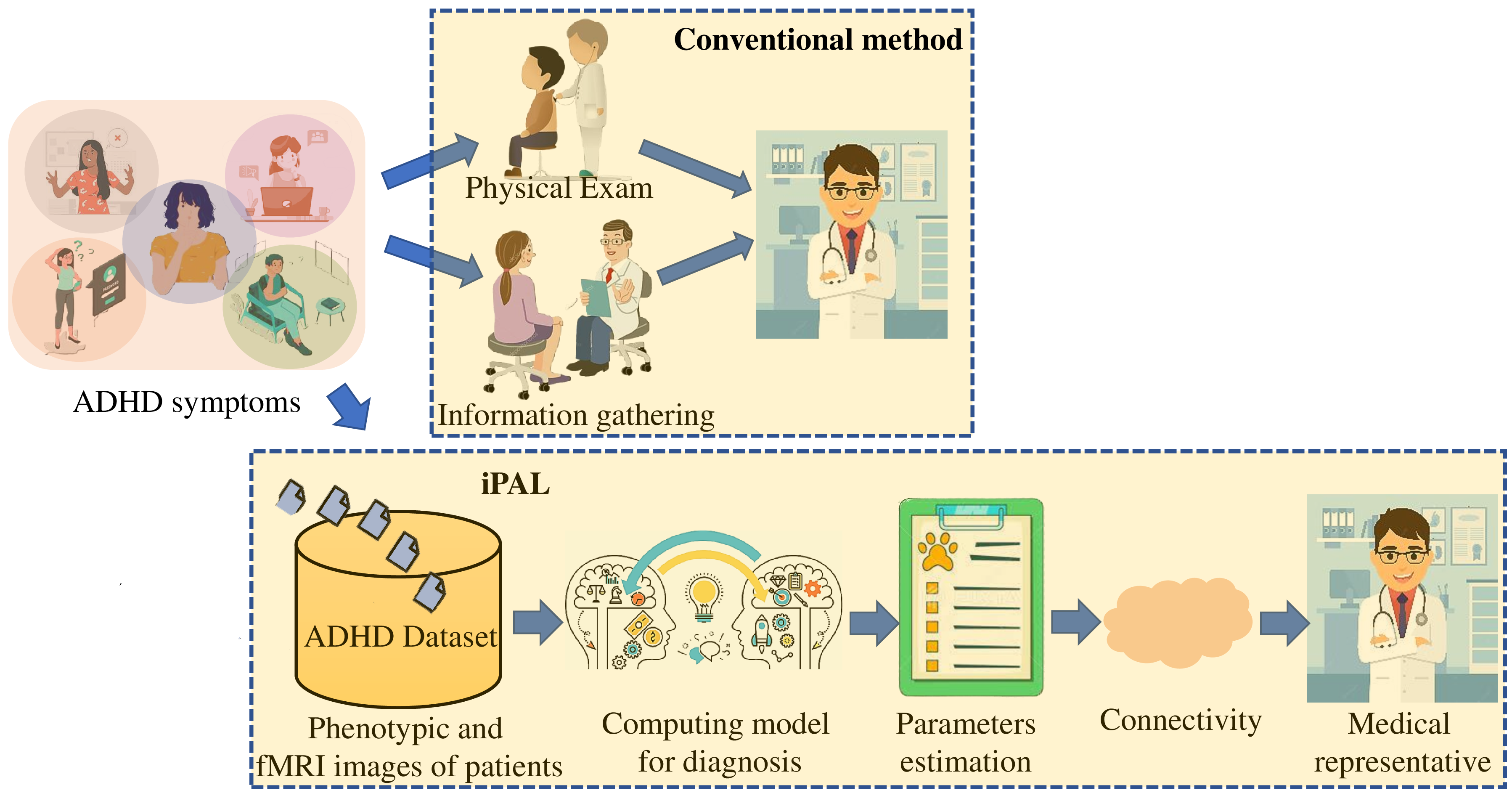}
	\DeclareGraphicsExtensions.
	\caption{An ADHD Diagnosis Framework for Smart Healthcare}
	\label{thematic}
\end{figure*}

To mitigate such kind of issues, an ADHDcare framework iPAL is proposed for smart healthcare. Based on symptoms, the brain images will be captured and processed through computing model, which will be already trained and tested through the samples. The results would be available in terms of status of disorders. The results status would be updated to medical experts, who is a available at remote location and connected through the internet servers. The medical experts would be available to the patients for counseling according to preferred time. The data would be available on cloud server, which can be accessed through user, medical staff and experts for treatment and clinical studies. The conventional process and proposed system is visualized in Figure \ref{thematic}.

\subsection{Novelty and Significance of the Proposed Solution}
 
\begin{figure*}[htbp]
	\centering
	\includegraphics[width=0.9\textwidth]{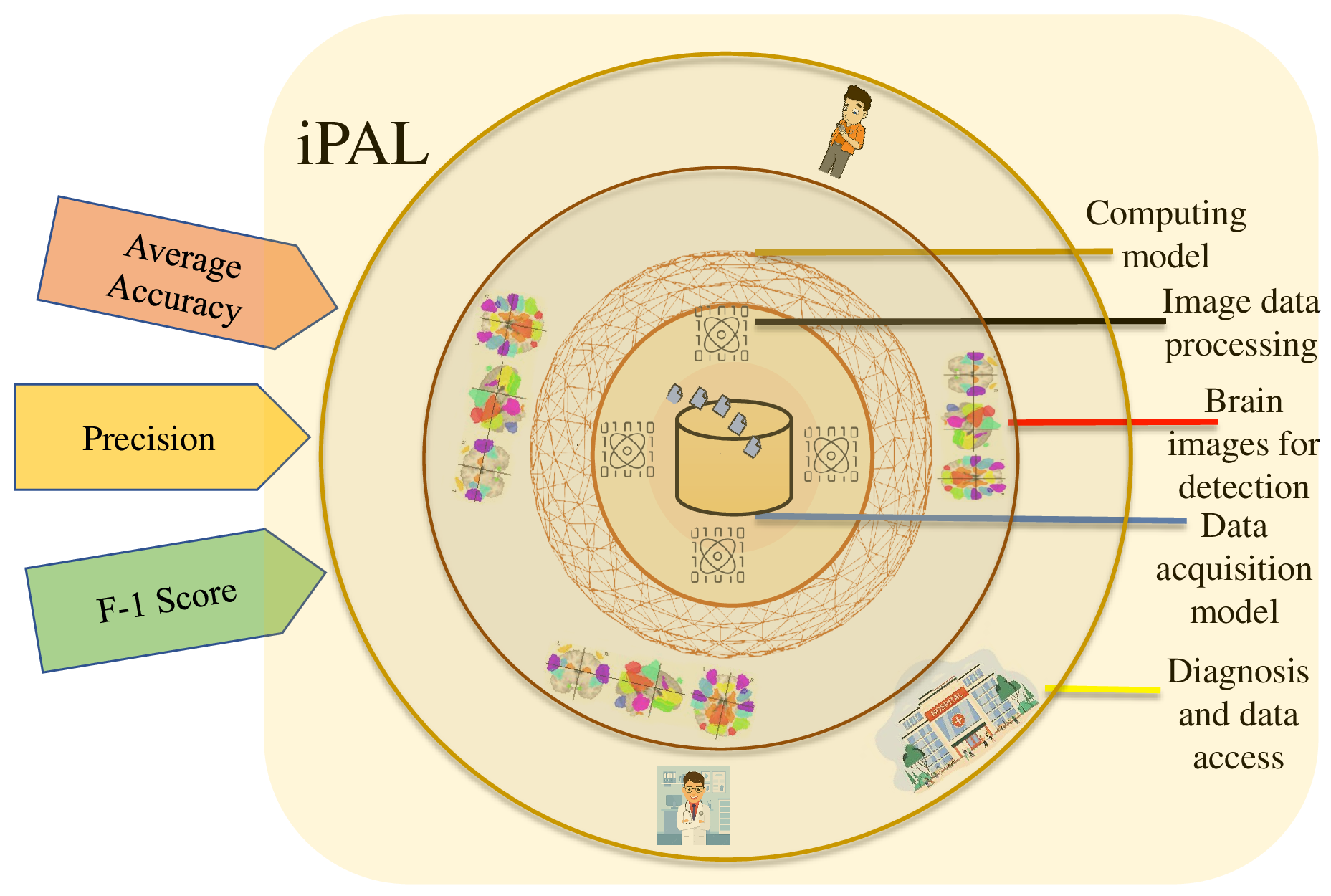}
	\DeclareGraphicsExtensions.
	\caption{Significance and effective paradigm of iPAL }
	\label{sign}
\end{figure*}

Till today, all the research work that has been done on precise level justifications of mental disorder, which is highlighted in earlier work. Many models classify the EEG signals and different kind of brain images. This framework has been calibrated and tested on significant ADHD-200 dataset with sufficient subjects with multiple conditions. Phenotypic and fMRI images are taken for preprocessing, conditioning and classifying steps for ADHD detection. Multiple features are extracted for classification. Data has been segregated as per the standards of the analysis. This framework is portrayed as a virtual effective model, which would be the real-time framework for instant ADHD diagnostic.
The research contributions are judicially reported, which depict the advancement in recent technology for ADHD detection
\begin{enumerate}
	\item A real-time framework with fMRI and phenotypic images has been proposed for smart healthcare, where the patients can approach to medical experts at remote location.
	\item Key-point features have been extracted and precise classifier has been used for ADHD diagnosis with prescribed protocols of data analysis. 
	\item Live brain images are used to train models and trained modules or data have been utilized among the people for precise detection.
	\item Multiple classifiers are trained and tested for analysis of optimized model, which provide reliable results for mental disorder justifications.
	\item Attractive performance parameters have been achieved from precise model for reliable diagnosis.
\end{enumerate} 
The proposed work presented a distinct methodology, which results a precise justification of mental disorder from live phenotypic and fMRI images. The visualization is presented in Figure \ref{sign}, which shows the significance of the proposed framework.

\section{Automatic ADHD Diagnosis Framework : An overview of iPAL}
\begin{figure*}[htbp]
	\centering
	\includegraphics[width=0.9\textwidth]{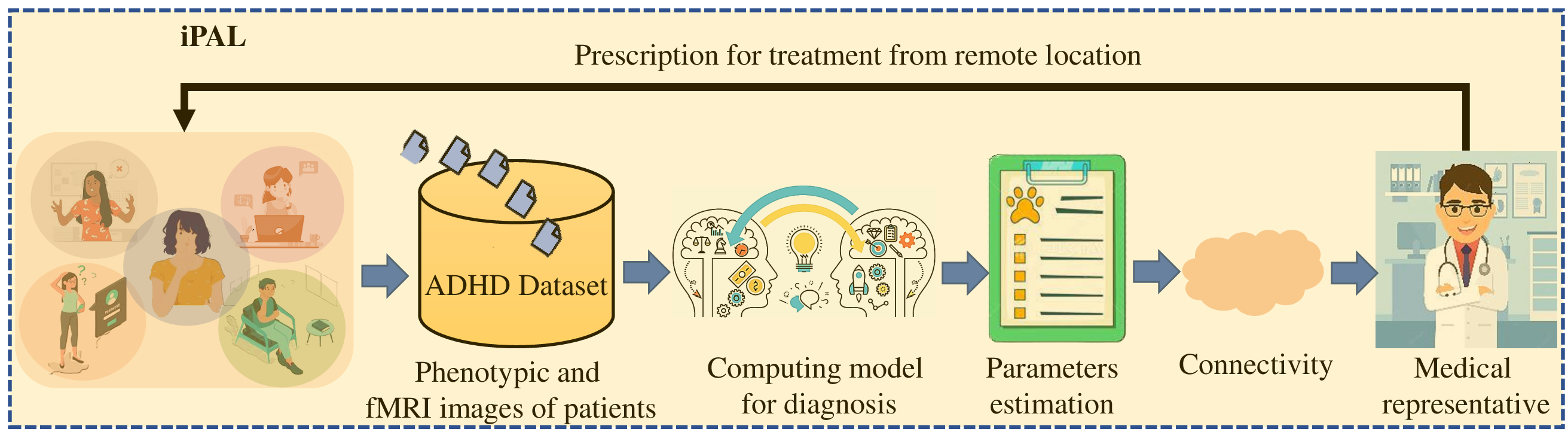}
	\DeclareGraphicsExtensions.
	\caption{An architectural Representation of iPAL}
	\label{arch}
\end{figure*}

In this presented work, different patients (subjects) have been taken to organize the efficient dataset to train the framework for ADHD diagnosis. Data has been segregated in 70-30\% ratio for training and testing. The live images are pre-processed and features are extracted for classification. The accuracy parameters have been calculated after the analysis through multiple models. The results are stored at cloud server for further access through the medical experts and patients. The data can be used and secured for further clinical trials through multiple health centers. The proposed iPAL would be able to diagnose the ADHD after the training and testing. The main advantage of the proposed framework is that this would be faster, reliable and low cost solution as there would be low cost to justify the mental disorder through iPAL and framework would be accessible through the patients and doctors at remote location. So, it will be much faster and reliable comparatively. The architectural view of proposed iPAL is represented in Figure \ref{arch}.

\subsection{Dataset for Proposed iPAL Framework Calibration and Initial testing}
\begin{figure*}[htbp]
	\centering
	\includegraphics[width=0.8\textwidth]{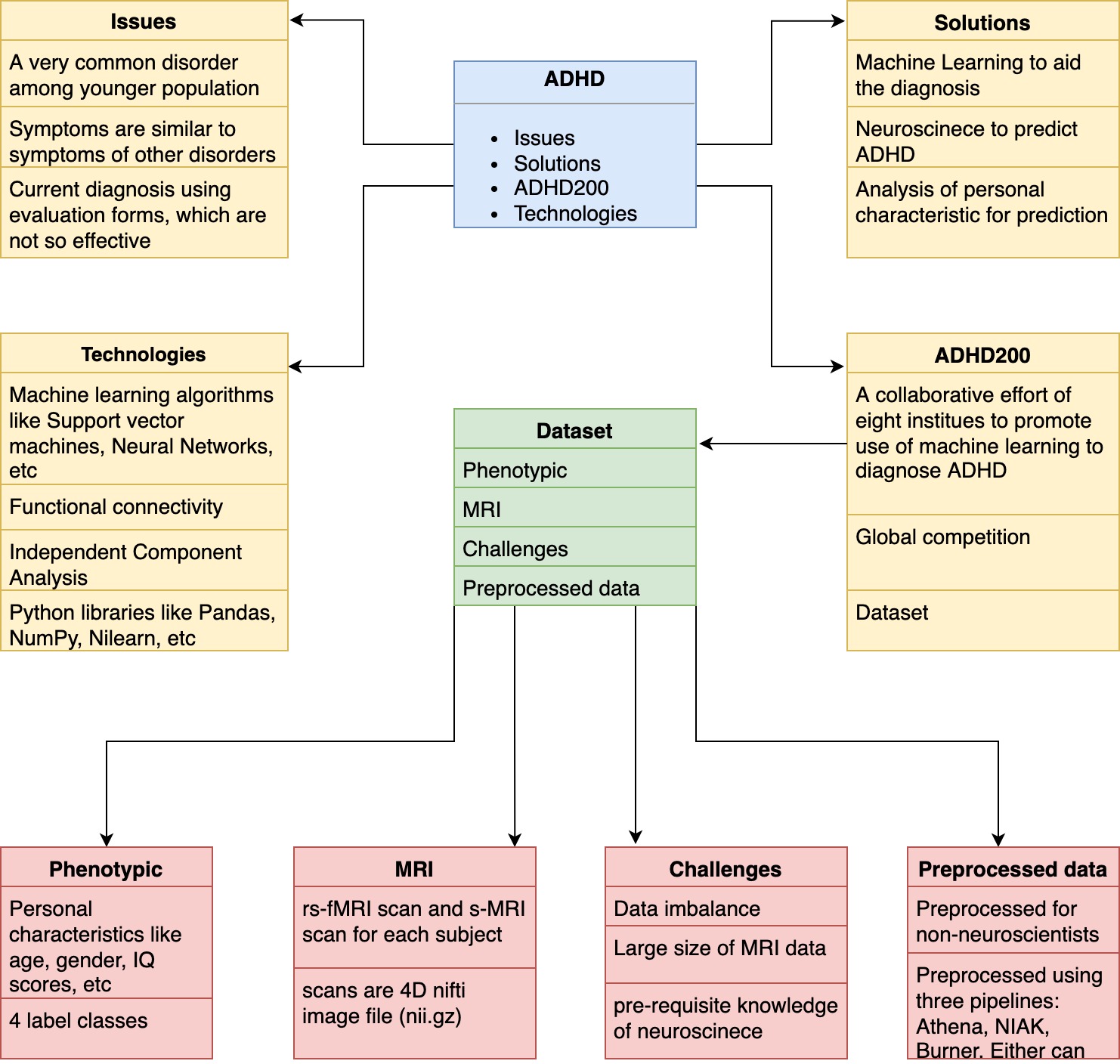}
	\DeclareGraphicsExtensions.
	\caption{ADHD dataset flow chart }
	\label{6}
\end{figure*}

As per the demand to solve the issues of mental disorder, a
According to prior work, overall ~2.2\%  of the average prevalence of ADHD has been estimated in children and adolescents (aged <18 years) [6]. Some common ways to diagnose ADHD are by evaluation forms, physical examinations, interviews with the patient, and similar methods. Hence, many studies have been conducted based on intelligent learning to diagnose ADHD effectively. For diagnosing ADHD, the ADHD-200 consortium globally held the ADHD-200 competition backed by the International Neuroimaging Data-sharing Initiative (INDI). Figure \ref{6} roughly depicts the motivation and purpose behind ADHD-200 and its dataset. The ADHD-200 dataset consists of resting-state functional-MRI(rs-fMRI) and structural magnetic resonance imaging (s-MRI) images of more than 900 subjects. The competition was focused on determining the prediction accuracy differentiating Typically Developing Children (TDC) and patients of ADHD. Initially, using imaging data, the highest accuracy was found to be 60.51\% in 2011. This motivated numeroud researchers to use data from the competition to perform various studies and predictions on ADHD, and many works are still going on to this date. The NITRC also made the preprocessed ADHD-200 data publicly available to facilitate the involvement of more researchers to develop further and provide better results.\\
The ADHD-200 dataset consists of two types of data, the personal characteristic data and the MRI (Magnetic Resonance Image) data consisting of a resting-state functional magnetic resonance image (rs-fMRI) and a structural magnetic resonance image(s-MRI). However, with the introduction of the ADHD-200 global competition, there has been a significant increase in works. The ADHD-200 data set facilitated these studies by providing a dataset of such scale.Therefore, it has seen the introduction of numerous models, and a lot are still developing.The Attention Deficit/Hyperactivity Disorder (ADHD) is a pervasive neurodevelopmental disorder affecting the younger population. However, ADHD does also affects numerous adults. People diagnosed with ADHD generally experience inattention (unable to focus properly), hyperactivity (excessive movement that is not fit for the sitting), and impulsive behaviours (may act without thinking about what the result will be). According to prior work, overall ~2.2\%  of the average prevalence of ADHD has been estimated in children and adolescents (aged <18 years) [6]. Some common ways to diagnose ADHD are by evaluation forms, physical examinations, interviews with the patient, and similar methods. Hence, many studies have been conducted based on intelligent learning to diagnose ADHD effectively. For diagnosing ADHD, the ADHD-200 consortium globally held the ADHD-200 competition backed by the International Neuroimaging Data-sharing Initiative (INDI). Figure \ref{6} roughly depicts the motivation and purpose behind ADHD-200 and its dataset. The ADHD-200 dataset consists of resting-state functional-MRI(rs-fMRI) and structural magnetic resonance imaging (s-MRI) images of more than 900 subjects. The competition was focused on determining the prediction accuracy differentiating Typically Developing Children (TDC) and patients of ADHD. Initially, using imaging data, the highest accuracy was found to be 60.51\% in 2011. This motivated numeroud researchers to use data from the competition to perform various studies and predictions on ADHD, and many works are still going on to this date. The NITRC also made the preprocessed ADHD-200 data publicly available to facilitate the involvement of more researchers to develop further and provide better results. The ADHD-200 dataset consists of two types of data, the personal characteristic data and the MRI (Magnetic Resonance Image) data consisting of a resting-state functional magnetic resonance image (rs-fMRI) and a structural magnetic resonance image(s-MRI). However, with the introduction of the ADHD-200 global competition, there has been a significant increase in works. The ADHD-200 data set facilitated these studies by providing a dataset of such scale.Therefore, it has seen the introduction of numerous models, and a lot are still developing.

Figure \ref{6} roughly depicts the motivation and purpose behind ADHD-200 and its dataset. The ADHD-200 dataset consists of resting-state functional-MRI(rs-fMRI) and structural magnetic resonance imaging (s-MRI) images of more than 900 subjects [10, 11]. The competition was focused on determining the prediction accuracy differentiating Typically Developing Children (TDC) and patients of ADHD. Initially, using imaging data, the highest accuracy was found to be 60.51\% in 2011. This motivated numeroud researchers to use data from the competition to perform various studies and predictions on ADHD, and many works are still going on to this date. The NITRC also made the preprocessed ADHD-200 data publicly available to facilitate the involvement of more researchers to develop further and provide better results. The ADHD-200 dataset consists of two types of data, the personal characteristic data and the MRI (Magnetic Resonance Image) data consisting of a resting-state functional magnetic resonance image (rs-fMRI) and a structural magnetic resonance image(s-MRI). However, with the introduction of the ADHD-200 global competition, there has been a significant increase in works. The ADHD-200 data set facilitated these studies by providing a dataset of such scale.Therefore, recent times have seen the introduction of numerous models, and a lot are still developing.
\subsection{Description of presented Dataset for automatic diagnosis of ADHD}
The Phenotypic Dataset consists of the personal characteristic data of the 973 participants from the ADHD-200 dataset, like ScanIDDir, Site, Gender, Age, Handedness, DX, Secondary DX, ADHD Measure, ADHD Index, Inattentive, Hyper/Impulsive, IQ Measure, Verbal IQ, Performance IQ, Full2 IQ, Full4 IQ, Med Status, QC\_Athena, QC\_NIAK, as mentioned in Table 1. We filtered out some rows to handle this imbalance since the dataset had some discrepancies and implicit values. For example, this data’s label, i.e., ‘DX’, had four values: ADHD-C, ADHD-H, ADHD-I, and Healthy Control/ Typically Developing Control (TDC), and was used in a multiclass classifier model.

For the MRI data, a subset of 824 subjects from the Athena Pipeline preprocessed data \cite{nitrc} was prepared using custom python scripts and used for the fMRI model, which included data from all eight websites. The data was filtered to use the preprocessed resting-state fMRI scans. This filtration amassed the relevant data of subjects at one place to be used, along with the phenotypic information of the individuals. This data was loaded using a custom python script to organise the data into a usable format.
This data’s label, i.e., ‘ADHD’/ ‘label’ originally also had four values: ADHD-C, ADHD-H, ADHD-I, and TDC, later converted to binary values, ‘1’ for ADHD and ‘0’ for not having ADHD.
A sample of 40 subjects provided by the nilearn library was also used.

The phenotypic data taken from the ADHD-200 competition from NITRC of 973 subjects was used to predict and diagnose ADHD in children aged 7-18. As discussed in [8], the proposed multi-objective classification scheme avoids hyper-parameter selection. Furthermore, the parameters were balanced to avoid overfitting. Phenotypic data consists of personal characteristics like ScanIDDir, Site, Gender, Age, Handedness, DX, Secondary DX, ADHD Measure, ADHD Index, Inattentive, Hyper/Impulsive, IQ Measure, Verbal IQ, Performance IQ, Full2 IQ, Full4 IQ, Med Status, QC\_Athena, QC\_NIAK (Table 1). The data consisted of four classification types, namely, typically developing children (TDC), ADHD-inattentive (ADHD-I), ADHD-Hyperactive/Impulsive(ADHD-H), and ADHD-combined (ADHD-C). The data had some implicit and missing values. Unnecessary columns were sorted out and rows missing crucial data for ADHD prediction were removed to handle these implicit values and discrepancies. Upon sorting, data size was reduced to 505 subjects and applied to various classification algorithms to better predict the data. The classification models mainly used are support vector machine, random forest, AdaBoost, KNN, logistic regression. Support vector machine predicted with the most accuracy along with the classification the graph plots between the parameters show significant relations between ADHD and possible factors affecting it.

This work integrates the use of phenotypic of the individuals with the resting-state functional-Magnetic Resonance Images. When performed on a sample of 40 subjects from the nilearn dataset, the Support Vector Machine provides an accuracy of 100\%. In contrast, the neural network implemented on functional connectome data achieves an accuracy of 75\%. Using a voting classifier with equal weightage to diagnose ADHD, an accuracy of 100\% was achieved. Such a combined high accuracy is because the neural network implemented on functional-MRI is 100\% accurate in predicting ADHD but not perfect for predicting Control. Hence, SVM can filter out the Control, resulting in overall accuracy is 100\%.
This method was further extended and applied to the data present in publicly available ADHD-200 dataset. With the Independent Component Analysis (ICA) [19], multiple Regions of Interest(ROIs) are extracted from the voxel time-series of the fMRI using the CanICA. These ROIs further help compute functional connectivities like 'tangent', 'partial correlation' and 'correlation'. The efficient functional connectivity obtained is additionally used to plot the functional-connectivity maps. According to [15], functional connectivity refers to undirected coupling strength between voxels or regions and is calculated using standard correlation measures. Thus, functional-connectivity maps are a representation of these connections. Therefore, functional connectivity can be further used as an input for the predictive models.

The model in prior work represented an accuracy of 65\% [20-27]. It was observed that accuracy of 92.50\% is achieved. So here, this work tries to improve the accuracy even better.

For this work, the database was extended with the inclusion of more patient records. In Total, 824 patient records were used from [11]. Further analysing the usable records due to discrepancies and implicit values, the records were narrowed down to a count of 400 after data processing. In this way, the learning component of the algorithm was extended, resulting in better generalisation of the model. The dataset with 400 subjects was further analysed with a Support Vector Machine(SVM) on its phenotypic data, and with a neural net on the function-MRI data, for classification. The SVM gave an accuracy of  99.16\% on the phenotypic, and the neural net on MRI-based data gave an accuracy of 86.66\%. Finally, on further classifying the data, data ensembling using voting with equal weightage improves accuracy, resulting in 92.50\% accuracy.
Presently the data uses a wide range of parameters to predict ADHD in children, possible symptoms, and age at which is the children more prone to ADHD.

\section{Data Characteristics and preprocessing}
\subsection{Phenotypic Dataset}

Datasets from [11] and [28] were used, including both phenotypic and MRI datasets. One of which is a dataset of 40 subjects taken from the nilearn library. It has four significant dataset types i.e. 'func’, 'confounds', 'phenotypic' and 'description'. The data contains many features to help diagnose ADHD in patients, and particularly using phenotypic as it contains features like f0, RestScan, MeanFD, NumFD\_greater\_than\_020, rootMeanSquareFD, FDquartiletop14thFD, PercentFD\_greater\_than\_020, Subject, MeanDVARS and more. In the data pre-processing, the features for which most of the values are not available or null are eliminated. After dropping all such features and optimally processing the data, the count is reduced to 39 subjects and 18 phenotypic features, namely f0, Subject, RestScan, MeanFD, NumFD\_greater\_than\_020, rootMeanSquareFD, FDquartiletop14thFD, PercentFD\_greater\_than\_020, MeanDVARS, MeanFD\_Jenkinson, site, data\_set, age, sex, TDC, ADHD, sess\_1\_rest\_1 and sess\_1\_anat\_1, as depicted in Table 1, to apply SVM.

\begin{figure*}[!t]
\centering
\includegraphics[width=0.8\textwidth]{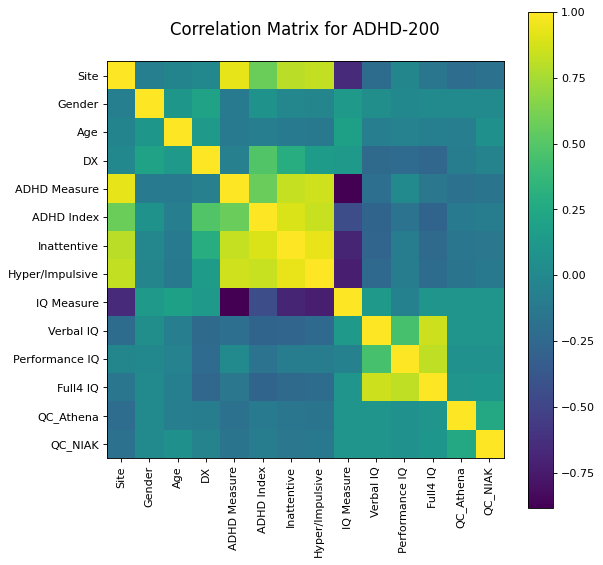}
 \DeclareGraphicsExtensions.
\caption{The correlation matrix depicts the relationship between multiple attributes of the phenotypic dataset used for classification.}
\label{7}

\end{figure*}

Another dataset is taken from [11], which contains the phenotypic and MRI data of the subject. The data was extracted for each subject and then concatenated to form a distinguished dataset. The phenotypic dataset was analysed further for 824 subjects and features, namely ScanDir ID, Site, Gender, Age, Handedness, ADHD, ADHD Measure, Inattentive, Hyper/Impulsive,  IQ Measure,   Verbal IQ,  Performance IQ, Full4 IQ, Med Status, QC\_Athena, and QC\_NIAK, leading to the elimination of the features containing more than 60\% null or not defined values, followed by dropping the remaining rows with null values from the dataset. After cleaning the data,  400 entries and 14 features, namely ScanDir ID, Site, Gender, Age, ADHD, ADHD Measure, Inattentive, Hyper/Impulsive,  IQ Measure,   Verbal IQ,  Performance IQ, Full4 IQ, QC\_Athena and QC\_NIAK, as depicted in Table 1, were left to be used for further analysis. The correlation matrix of multiple attributes for ADHD-200 dataset is represented in Figure \ref{7}.
\begin{figure*}[!t]
\centering
\subfigure{\includegraphics[width=0.8\textwidth]{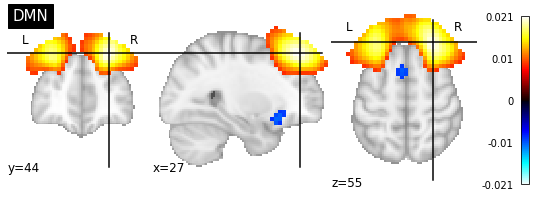}
}
\subfigure{\includegraphics[width=0.8\textwidth]{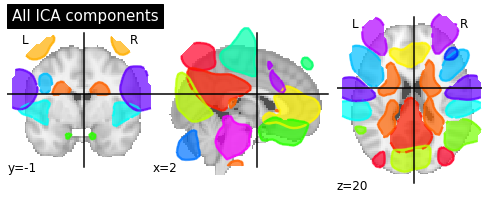}%
}
\caption{ Visualization of the Default Mode Networks(DMN) and the Components obtained from the Independent Component Analysis(ICA) on a 3D brain mask image using Nift Map Masker.}
\label{8}
\end{figure*}
Visualization of the Default Mode Networks(DMN) and the Components obtained from the Independent Component Analysis(ICA) on a 3D brain mask image using Nift Map Masker is enlighten by Figure \ref{8}.

\begin{figure*}[!t]
\centering
\includegraphics[width=0.8\textwidth]{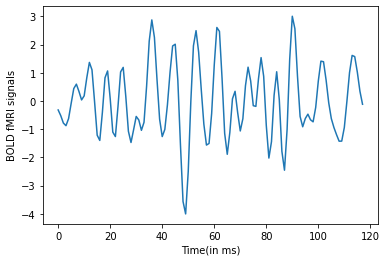}
 \DeclareGraphicsExtensions.
\caption{Time-series visualization of a voxel of an MRI scan from ADHD-200 sample.}
\label{9}
\end{figure*}
\begin{figure*}[!t]
\centering
\includegraphics[width=0.8\textwidth]{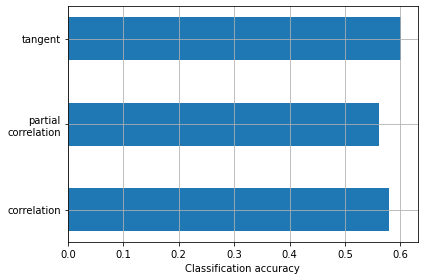}
 \DeclareGraphicsExtensions.
\caption{Accuracy plot of connectivity biomarkers: ‘tangent’, ‘partial correlation’, ‘correlation’}
\label{10}
\end{figure*}

Time-series visualization of a voxel of an MRI scan from ADHD-200 sample. Here the x-axis corresponds to the time units, and the y-axis corresponds to the blood-oxygen-level-dependent (BOLD) fMRI signal.  Voxels can be classified into ‘active’ or ‘inactive’ based on the intensity of these signal. This time series, on further computation, results in functional connectivity. It is visualized in Figure \ref{9}. Accuracy plot of connectivity biomarkers is shown in Figure \ref{10}.

The phenotypic file is chosen from [11] to work for the diagnosis and classification of ADHD. This file contains data for 973 individuals with many discrepancies, and hence it was needed to handle this imbalance before feeding this data to multiple multiclass classification models. Various permutations of features were adopted to determine the result, keeping in mind the usefulness of those features regarding ADHD, narrowing down to 505 individuals with multiple features. These include Site, Gender, Age, ADHD Measure, ADHD Index, Inattentive, Hyper/Impulsive, IQ Measure, Verbal IQ, Performance IQ, Full4 IQ, QC\_Athena, and QC\_NIAK, 
as depicted in Table \ref{5}. The correlation matrix in Figure \ref{7} shows the relationship between multiple features, and the column distribution helps understand the spread of elements in the data.

\subsection{MRI Dataset}
The MRI data used is taken from the Athena pipeline preprocessed data from [11]. In total, 824 subjects' data was used, and after handling the implicit data, it was narrowed down to 400. The rs\-fMRI data was in the form of a 4D nifti image file(.nii.gz) with noise variables removed. In a 4D nifti image, the first three dimensions represent the three coordinates of a 3D space, i.e., x,y, and z, and the 4th dimension represents time. Thus, a 4D image is a collection of several 3D images corresponding to each unit of time. A 3D image represents voxels, just like a 2D image represents pixels. Each voxel has some fMRI signal associated with it, which is a blood-oxygen-level-dependent (BOLD) signal (detected in fMRI) in this case. Plotting these signals of a voxel against the time is called a time series. These time series provide further functional connectivity computation data, potentially allowing machine learning applications and research. The proper notations for ADHD-200 phenotypic are represented in Table \ref{5}.

\begin{figure*}[!t]
\centering
\subfigure[first case]{\includegraphics[width=0.6\textwidth]{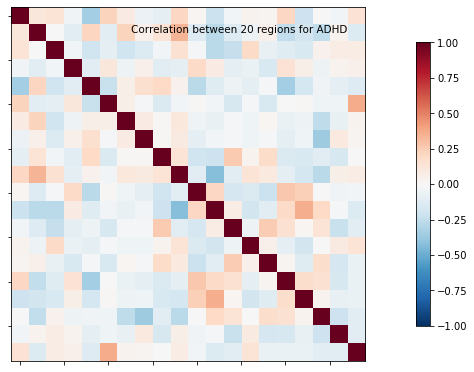}%
}
\hfil
\subfigure[second case]{\includegraphics[width=0.6\textwidth]{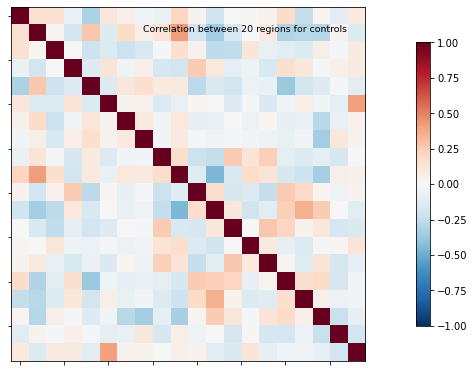}%
}
\caption{Correlation matrix of mean functional connectivity of all the \textcolor{blue}{ADHD and TDC} subjects}
\label{12}
\end{figure*}

\begin{table*}[htbp]
\renewcommand{\arraystretch}{1.3}

\caption{Notations related to features used in the ADHD-200 phenotypic}
\label{5}
\centering
\begin{tabular}{ |p{3.95cm}||p{4cm}||p{8.3cm}|}
\hline
 parameter & description & values \\ \hline
Subject ID/ an individual ID & Seven digits integral value &\\
\hline
  Site & to provide the individual’s data &  1 Peking University 2   Bradley Hospital/Brown University 3 Kennedy Krieger Institute 4 NeuroIMAGE Sample 5 New York University Child Study Center 6 Oregon Health \&  Science University 7 University of Pittsburgh 8 Washington University in St. Louis \\
\hline
Gender/sex & Gender information  & 0 Female 1 Male \\ \hline
 Age & Age at the time of test & Children aged 7\- 18. Mean: 11.37 years \\ \hline
DX & Diagnosis of ADHD & 0 - Typically Developing Children
1 - ADHD - Combined
2 - ADHD - Hyperactive/Impulsive
3 - ADHD - Inattentive \\ \hline
ADHD & Diagnosis of ADHD & 0 - Typically Developing Children
1 - ADHD - Combined, Hyperactive/Impulsive, Inattentive \\ \hline
ADHD Measure & measuring method & 1 ADHD Rating Scale-IV (ADHD-RS)
2 Conners’ Parent Rating Scale-Revised, Long version (CPRS-LV)
3 Connors’ Rating Scale-3rd Edition \\ \hline
ADHD Index & ADHD measurement & Values between 18 - 99. Mean: 49.37 \\ \hline
Inattentive value & based on ADHD measure adopted & Values between 9 - 90. Mean: 41.61 \\ \hline
Hyper/Impulsive value & ADHD measure adopted &
Values between 9 - 99 Mean: 39.85 \\ \hline
IQ Measure & The method used for measuring IQ & 1 Wechsler Intelligence Scale for Children, Fourth Edition (WISC-IV) 2 Wechsler Abbreviated Scale of Intelligence (WASI) 
3 Wechsler Intelligence Scale for Chinese Children-Revised (WISCC-R) 
4 Two subtest WASI 
5 Two subtest WISC or WAIS – Block Design and Vocabulary \\ \hline
Verbal IQ & based on IQ measure & Values between 65 - 158. Mean: 112.80 \\ \hline
Performance IQ & based on IQ measure & Values between 54 - 139. Mean: 106.01 \\ \hline
Full4 IQ &  Full4 IQ measured &
Values between 73 - 153. Mean: 110.41 \\ \hline

QC\_Athena & Quality Control check of an individual’s data wrt Athena pipeline & 
0 Questionable 
1 Pass
QC\_NIAK \\ \hline
Quality COntrol & check of an individual’s data wrt NIAK pipeline &
0 Questionable 
1 Pass \\ \hline
Rest Scan  & Type of resting-state scan & All are ‘rest\_1’ \\ \hline
MeanFD & Mean Functional Dependency & Float value between 0 to 1 \\ \hline
NumFD\_greater\_than\_020  & Functional Dependencies with strength > 20\% &
Integral value \\ \hline

rootMeanSquareFD & RMS of Functional Dependencies &
Float value between 0 to 1 \\ \hline
PercentFD\_greater\_than\_020 & Functional Dependencies with strength > 20\% (\%) & Float value between 0 and 1 \\ \hline
MeanDVARS  & Mean of the change in average whole\-brain signal &
Float Value \\ \hline
MeanFD\_Jenkinson & Mean correlation between motion parameter &
Float Value \\ \hline
data\_set & Type of dataset, initially used by nilearn &
‘test\_set’/’data\_set’ \\ \hline
tdc/ TDC & Abbreviation of Typically Developing Control &
0 or 1 \\ \hline

\end{tabular}
\end{table*}
\begin{figure*}[!h]
	\centering
	\subfigure[first case]{\includegraphics[width=0.6\textwidth]{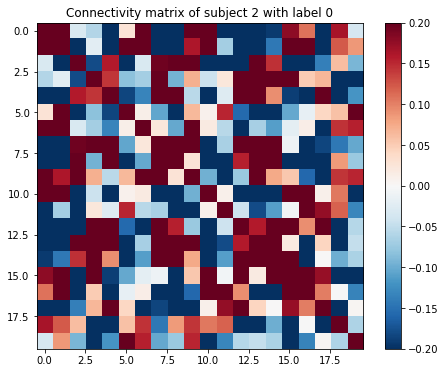}}\label{}
	\subfigure[second case] {\includegraphics[width=0.6\textwidth]{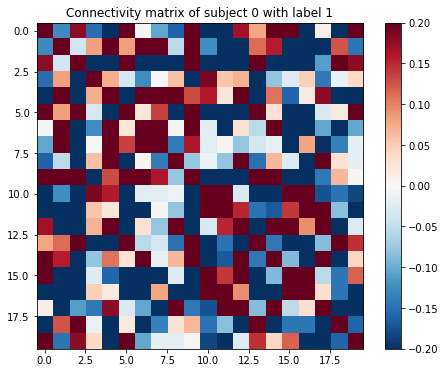}}\label{}
	\caption{Connectivity matrix between the 20 regions of a subject diagnosed with ADHD (label 1) with strength in the range -20\% to 20\%} 
	\label{11}
\end{figure*}
The connectivity matrix between 20 regions of a subject with ADHD and all of all subjects for mean functional connectivity are represented in Figure \ref{11} and \ref{12} respectively.

\subsection{Feature selection for ADHD Framework}
\subsubsection{Phenotypic dataset}
The Phenotypic dataset for 973 subjects contains multiple features, but not all contribute equally to the diagnosis of ADHD. By applying logistic regression and examining the coefficients, the features which contribute significantly can be understood. Such features include Site, Gender, ADHD Measure, ADHD Index,  Inattentive, Hyper/Impulsive, IQ Measure, Verbal IQ, and Performance IQ. Other features that do not affect significantly are  Age, Handedness, Full4 IQ, Full2 IQ, Med Status, QC\_Athena, and QC\_NIAK, and could be due to multiple factors. The data sample already represents a small fraction of age, and hence the diagnosis may be independent of an individual’s age in that range. Measurement of handedness depended on the University adopted for the measurement technique and consequently had large diversity. Some individuals had binary or ternary integral values, and others had a decimal value of handedness, leading to considerable inconsistency.  Med Status and Full2 IQ had almost no impact on the diagnosis and had immense implicit value, and hence, it was better to be removed to work upon a larger dataset. This was concluded after running the scripts with and without considering these attributes as well as finding the majority of the instances with these attributes having either no value available (N/A) or out-of-bounds values like -999. Therefore, one can not claim the usefulness of a feature beyond the scope of the dataset due to significant inconsistency. As when we sorted the data considering med status and Full2 IQ and passed it to the classifier there was no significant impact on result due to both of them   So, We have integrated phenotypic information of Site, Gender, Age, DX, ADHD Measure, ADHD Index, Inattentive, Hyper/Impulsive,  IQ Measure,   Verbal IQ,  Performance IQ, Full4 IQ, QC\_Athena, and QC\_NIAK, as depicted in Table 1, which are left to be used for further analysis with data for 505 subjects.
\subsubsection{MRI dataset}
The resting-state fMRI data had some fMRI files with their noise variables already removed and some still with noise variables present in them. Though time courses are a popular choice in neuroimaging, taking all the voxels and their time series can increase data size by a significant amount. Therefore the probabilistic and statistical analysis is performed to extract the ROIs, i.e., voxels which are temporarily correlated and are active. The rest of the voxels are purposeless and treated as noisy voxels. One such method is Independent Component Analysis (ICA). Using Independent Component Analysis, 20 components were extracted, as shown in Figure \ref{8}(b). Numerous computations ran, keeping the ROIs in the range of 15-40. The analysis yielded 20 as the optimum count.

Therefore Independent Component Analysis was performed using the group-level ICA (CanICA) method from [28] to analyse the fMRI signals and remove these noise variables. According to [29], Resting MRI signal(BOLD signals) processing has been crucial in identifying ROIs. Default Mode Networks (DMNs) effectively indicate the difference between an ADHD brain and a typically developing brain, and with ICA, ROIs are identified efficiently from the MRI [28-30]. By identifying the ROIs, the number of voxels can be significantly reduced to only activated voxels to plot the functional connectivity maps [27-29]. Removing the inactive/noisy voxels also improves accuracy. Figure \ref{8}(a) shows the default mode network for the components extracted using ICA plotted for the coordinate values ( x = 27, y = 44, z = 55), represented by the three images, each along with one of the axes and the point of intersection of lines shows the coordinate. Similarly, Figure \ref{8}(b) is a visualization of the 20 components extracted from performing the ICA of the dataset, taking (x = 2, y = -1, z = 20) as the reference coordinate.

After removing the noisy voxels and extracting the ROIs, the time-series of this was obtained data using the fit\_transform method of [28] like that shown in Figure \ref{9}. This time series, on further computation, can be used to obtain functional connectivity.

Three connectivity measures were considered to calculate the functional connectivity: correlation, partial correlation, and tangent. First, the accuracies for these three measures were comapred and the ‘correlation’ and ‘tangent’ were found to be similar and accurate enough, as shown in Figure \ref{10}, and proceeded with ‘correlation’. The results of [30-33] shows the correlation as a preference as a connectivity measure over others. Furthermore, the networks created using correlation as a connectivity measure appear extensively clustered compared with random networks. For this, the data was startified using StratifiedKFold cross-validation from the sklearn [34] library and classified using connectivity measures and Linear Support Vector Classifier for comparison.
\begin{figure*}[htbp]
\centering
\subfigure[first case]{\includegraphics[width=0.6\textwidth]{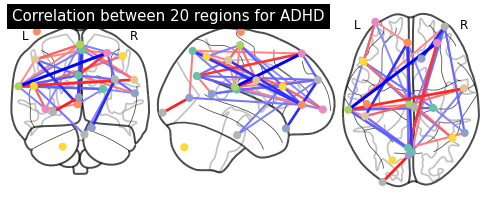}%
}
%\hfil
\subfigure[second case]{\includegraphics[width=0.6\textwidth]{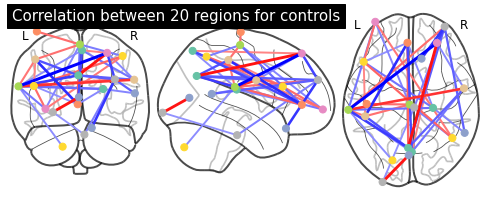}%
}
\caption{Mean Functional connectome for ADHD and TDC subjects as represented in a three dimensional space. In this figure, the color of the line represents the intensity of the connection(Refer to Figure \ref{14}).}
\label{13}
\end{figure*}
\begin{figure*}[htbp]
\centering
\subfigure[first case]{\includegraphics[width=0.4\textwidth]{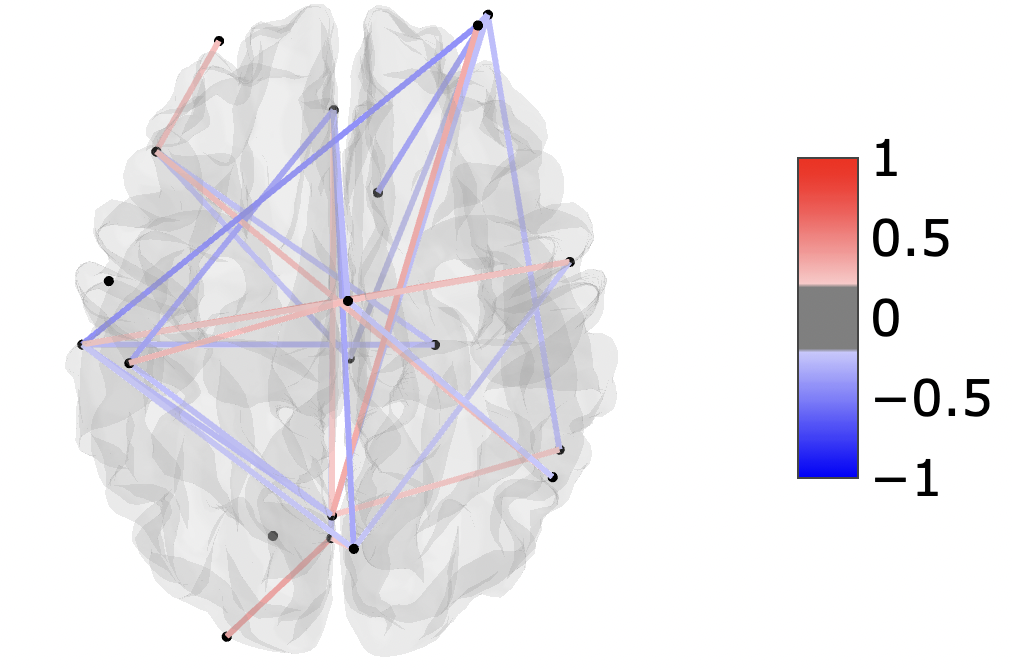}}%
\label{}
\hfil
\subfigure[second case]{\includegraphics[width=0.4\textwidth]{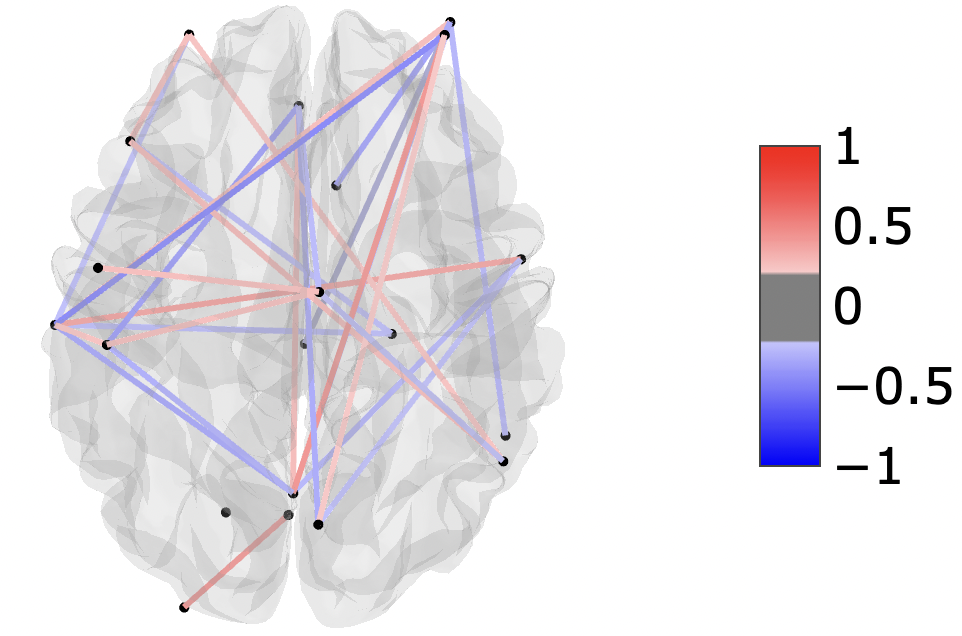}%
}\label{}
\caption{Functional connectivity map of the mean connectivity matrix showing the strength of connections in between 20 regions on a color scale and giving a better perception of density of these connections for ADHD and TDC subjects.}
\label{14}
\end{figure*}

After obtaining the time-series of the \textcolor{blue}{ROIs}, the functional connectivities are measured using the correlation connectivity measure in the form of correlation matrices. According to the work, functional connectivity refers to undirected coupling strength between voxels or regions, usually calculated using the standard correlation measures [27-29]. Functional connectivity maps are a representation of these connections. Further, functional connectivity can be used as an input to predictive models. Figure \ref{11} (a) and (b) shows the connectivity matrix for an individual without ADHD and an individual with ADHD, respectively.

Figure \ref{12} (a) and (b) are the mean correlation matrices of the mean functional connectivity for ADHD and TDC, respectively, for the 20 regions obtained from the Independent Component Analysis. After ignoring the diagonal(correlation of a region with itself is always 1), the connections do not seem too different by looking. It is difficult to distinguish between both figures visually. Therefore these connections are plotted on the brain as a functional connectivity map and visualised as shown in Figure \ref{13}.

The mapping of these functional connectivity matrices onto a brain image where each node represents an \textcolor{blue}{ROI} and edges represent the time-synchronized connectivity between the \textcolor{blue}{ROIs} with edge strength of more than 80\% to obtain the functional connectivity maps or functional connectome. Figure \ref{13}(a) shows the mean connectivity map of ADHD-diagnosed patients. It has 20 \textcolor{blue}{ROIs}. Figure \ref{13}(b) shows the strength of connections between these \textcolor{blue}{ROIs} with a scale for context.

From Figure \ref{13} (a) \& (b) and Figure \ref{14} (a) \& (b), the visual difference can be seen between ADHD and TDC connectivity map. It is easy to see that the ADHD connections are significantly less dense than the TDC connections. Hence, this also suggests that it is possible to differentiate between ADHD and TDC using functional connectivity and use functional connectivity as an input to the classifiers to predict ADHD. Moreover, some studies prove that brain connectivity is a biomarker of ADHD diagnosis [35,36].
\begin{figure*}[htbp]
\centering
\includegraphics[width=0.99\textwidth]{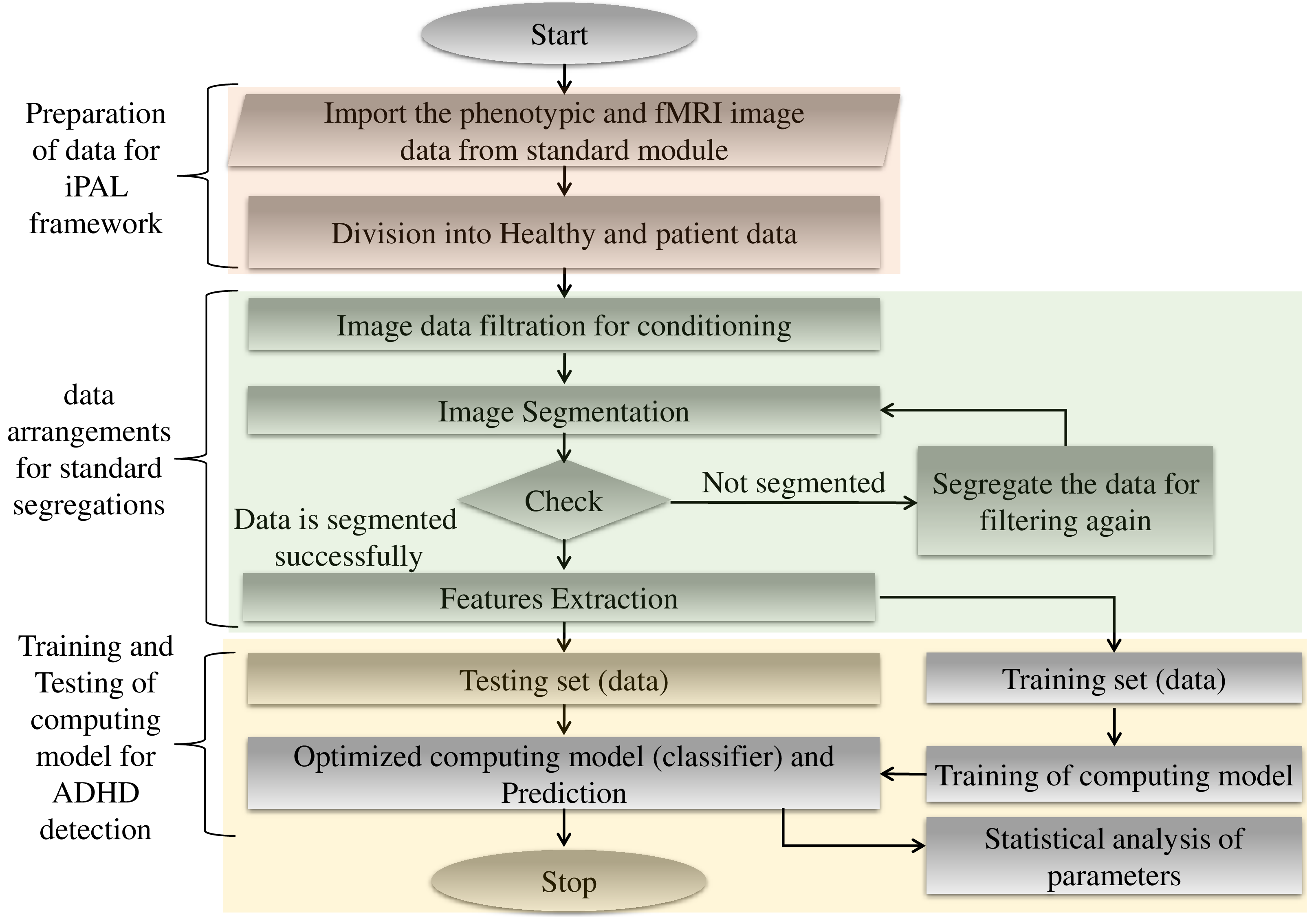}
 \DeclareGraphicsExtensions.
\caption{Processing Steps of an Automated ADHD diagnosis framework}
\label{15}
\end{figure*}

\begin{figure*}[htbp]
\centering
\includegraphics[width=0.8\textwidth]{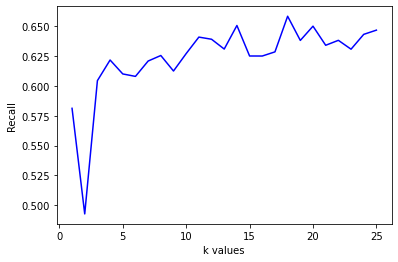}
 \DeclareGraphicsExtensions.
\caption{Graph of K parameter used in KNN}
\label{16}
\end{figure*}

\subsection{Classification for ADHD Framework}
\subsubsection{Phenotypic dataset}
An immense number of classifiers, such as K-nearest neighbour, Logistic Regression, Support Vector Machine, Artificial Neural Network, Random Forest, Ada-Boost, and Decision Tree, are used on the phenotypic dataset widespread and supervised learning classification methods. Supervised learning is where the classification results are known, and the models obtained by learning from these samples are used for prediction. Logistic Regression is a popular supervised classification algorithm that assigns weightage to each feature used to predict an individual’s class. This model is great when the dataset is linearly separable, and hence, is not best suited for our phenotypic dataset. The measuring techniques differ widely in the sites from which the data is gathered and the measuring techniques used by that side for measuring ADHD [35, 36].
Adaptive boosting (AdaBoost) is one of the most popular reinforcement algorithms as a supervised learning method. It combines weak classifiers with specific rules to build a robust classifier. AdaBoost determines the weight of each sample according to the last overall classification in each training session and the accuracy obtained.K-Nearest Neighbour is a supervised machine learning algorithm that is also used for classification problems. KNN, as the name suggests, that similar objects exist in close proximity to every other. First, the gap distance is calculated between the current point and the selected point in the algorithm. The distances are then added and sorted in ascending order, and finally, top K entries are chosen as nearest neighbours and are further used for determining the class of the current point.  Support vector machine is employed as described in [37] it gives the most effective accuracy among all the similar types of classifiers used. The SVM algorithm detects a hyperplane in an N-dimensional space that classifieds data points.Artificial Neural Network(ANN) is an excellent replacement for the SVM as its accuracy is very close to SVM. Its multilayered neural network is suitable for such a diverse dataset. Figure \ref{15} depicts the overall techniques and procedures used to process the datasets from different sites.
\subsubsection{MRI dataset}
The MRI dataset obtained as the functional connectivity biomarkers are further used to diagnose ADHD. As observed by plotting the correlation matrices and comparing them for ADHD and TDC, they seem very similar visually. Hence, a comprehensive evaluation was required to distinguish between individuals with ADHD and TDC. The dataset was observed to be linearly inseparable, independent of nearest neighbours, and did not follow a decision tree-based pattern. Using a neural network for classification appeared to be a potential candidate and was later observed to be the best suitable method.
\begin{figure}[htbp]
\centering
\includegraphics[width=0.5\textwidth]{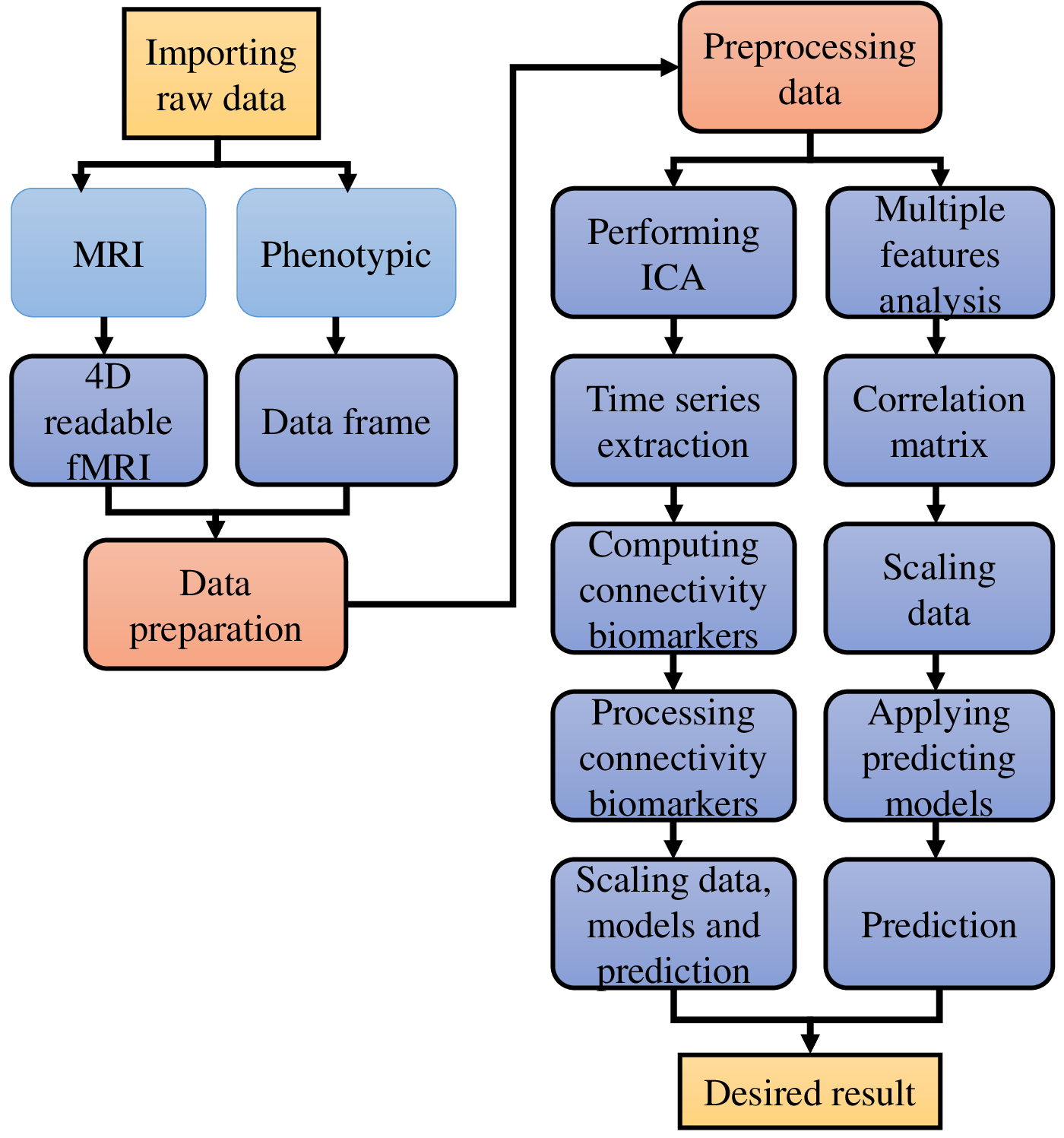}
 \DeclareGraphicsExtensions.
\caption{Depiction of the methodology used to obtain results }
\label{17}
\end{figure}
\subsection{Sophisticated Model training for Precise Diagnosis}
\subsubsection{Phenotypic dataset}

Initially, the phenotypic dataset taken from [11] was distributed into classification and prediction features. The classification features were then subjected to the transformation using a Min-Max Scaler. Further, the data of 505 subjects were split into train and test with a split size of 20\%. These 404 training subjects were processed under numerous classification algorithms, and the models obtained were used for prediction on 101 subjects. Multiple classification algorithms were applied to determine the best possible classifier for the dataset, including Logistic Regression, Random Forest, Ada-Boost, Support Vector Machine, k-Nearest Neighbour, and Neural Network.
Linear Regression uses cross-entropy loss function as the multi\_class is multinomial. Some of the parameters used include the solver as ‘lbfgs’, tolerance as 1e-4, and max iterations to be 101.
Grid Search was performed to determine the best parameters for the Random Forest, which result in ‘criterion’ as ‘entropy’, ‘max\_features’ as ‘sqrt’, and n\_estimators as 1000.Grid Search was also performed for SVM for best parameter estimation, resulting in regularisation parameter as 100, gamma as 0.04, and kernel as ‘rbf’.KNN was performed for the K parameter ranging from 1 to 26, and the following graph was observed in Figure \ref{16}.

Hence, the optimum value of K turned out to be 18, which contained numerous points from multiple classes, therefore eliminating KNN as a reliable classifier.While working with the ANN, the prediction class for training was converted using One-Hot-Encoder to be used in a neural network. Further, a Sequential Neural Network was created, having an input size of thirteen dimensions. Five dense layers were added with dimension and activation functions as eleven and ‘relu, nine and ‘relu’, seven and ‘relu’, five and ‘relu’, and lastly, four and ‘sigmoid’. The model was compiled using loss function as ‘categorical\_crossentropy’, optimiser as adam, and metric of accuracy. After training the neural network model with a max epoch count of one hundred, the prediction was made, and the result was converted into an array back from One-Hot-Encoded format.
\subsubsection{MRI dataset}
Since the combined model has to work on functional MRI and phenotypic individually, it is crucial to use the same individuals for training-testing in Neural Networks and the SVM. Therefore, the features and the prediction class for both functional MRI and phenotypic are zipped together and then split into train-test with 70\% training and 30\% testing, further divided into features and prediction class of both the fMRI data and the phenotypic data. However, before these training and testing data can be combined, they also need to be scaled to train with the model. While working with the ANN for fMRI, the prediction class was converted using One-Hot-Encoder to be used in a neural network. For both the fMRI and the phenotypic, the features were processed through the Min-Max Scaler separately. Another thing to note is that the diagnosis of ADHD is converted into binary form, whereas the phenotypic file had ADHD of four types, as mentioned as DX in Table 1.

\begin{figure*}[htbp]
\centering
\includegraphics[width=0.8\textwidth]{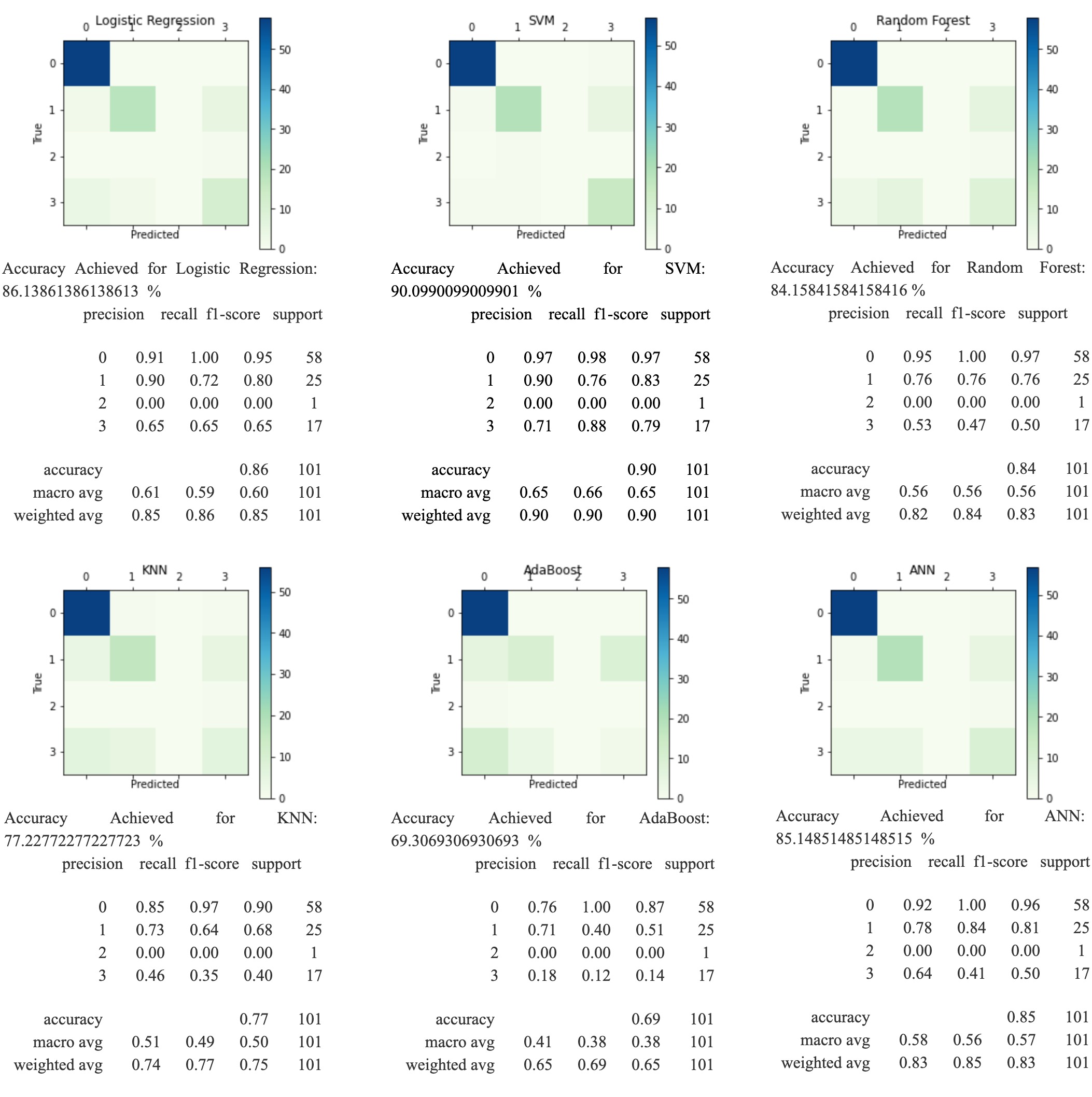}%
 \DeclareGraphicsExtensions.
\caption{Confusion matrix of all models used on Phenotypic data }
\label{18}
\end{figure*}

\begin{figure*}[htbp]
\centering
\includegraphics[width=0.8\textwidth]{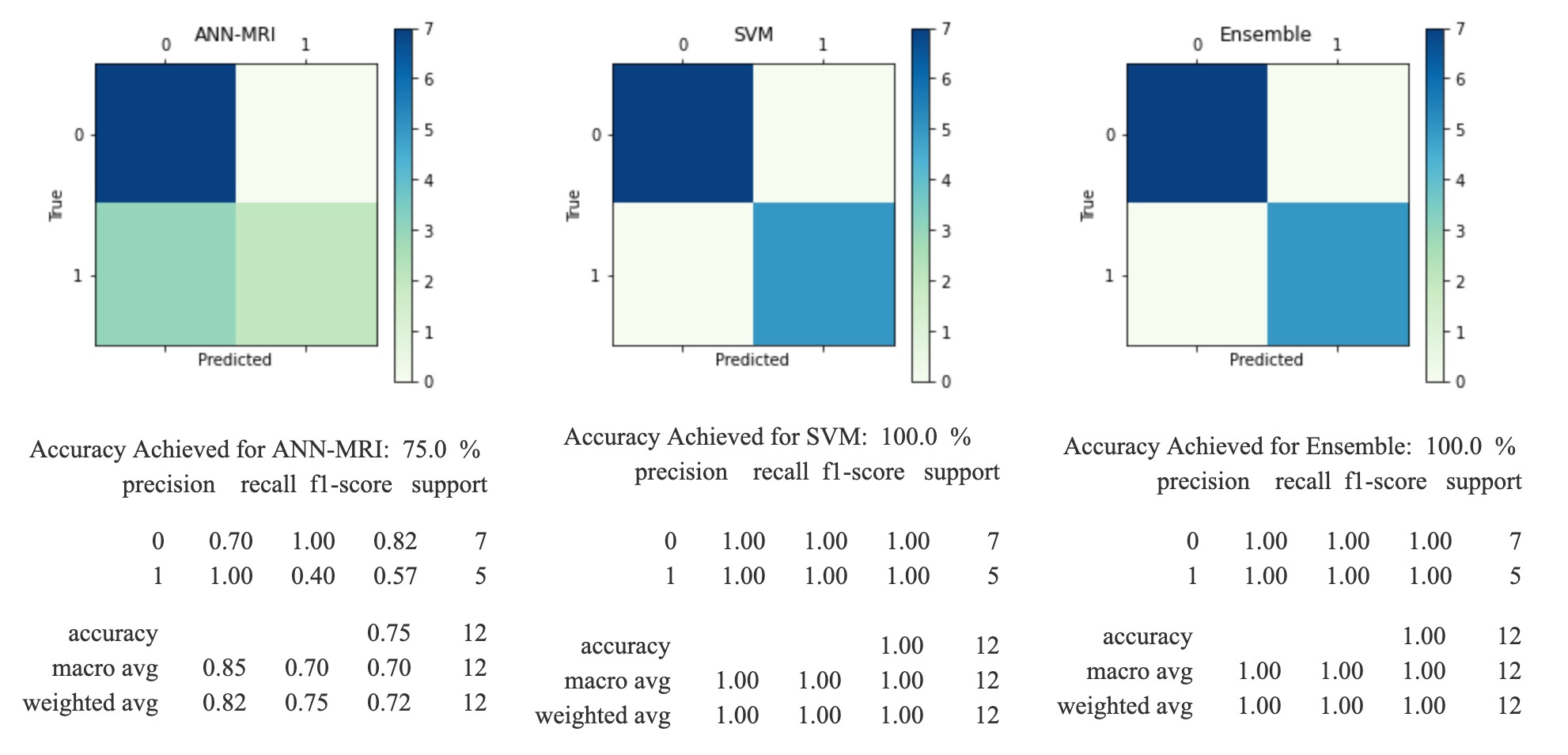}%
 \DeclareGraphicsExtensions.
\caption{Confusion matrix of the algorithm applied on nilearn dataset }
\label{19}
\end{figure*}

\begin{figure*}[htbp]
\centering
\includegraphics[width=0.8\textwidth]{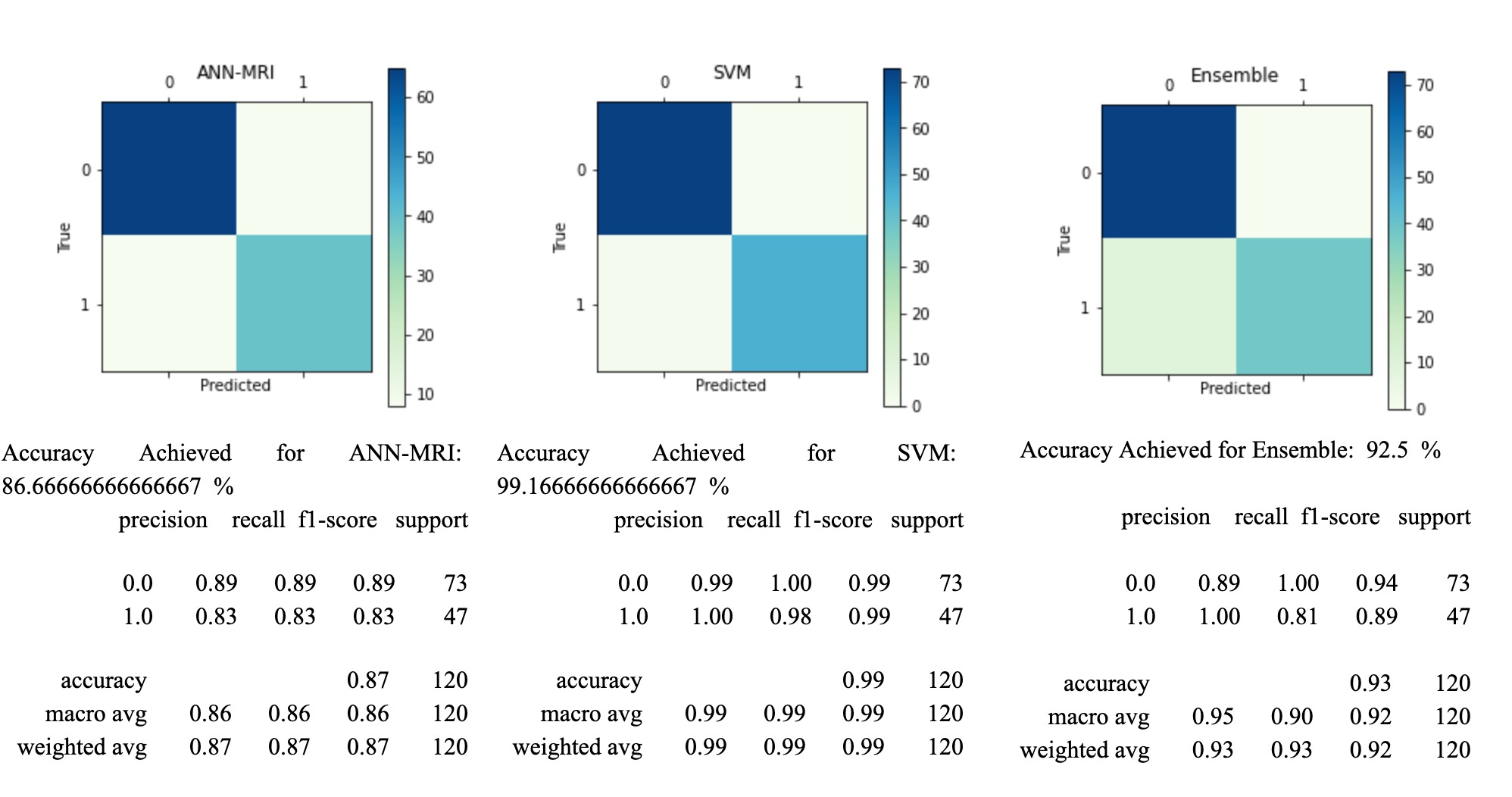}%
 \DeclareGraphicsExtensions.
\caption{Confusion matrix of the algorithm used on nitrc dataset }
\label{20}
\end{figure*}
\begin{figure*}[htbp]
\centering
\includegraphics[width=0.8\textwidth]{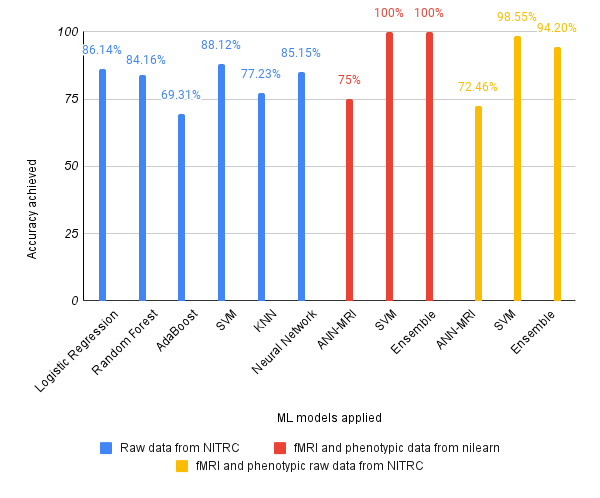} 

 \DeclareGraphicsExtensions.\caption{Comparison of the accuracy achieved for different ML models applied on different datasets }
\label{21}
\end{figure*}

The correlation matrix developed using functional connectivity as the connectivity biomarker is used as input data for our neural network. A Sequential Neural Network model made up of Dense layers was built, having an input size of 210 neurons. Seven dense layers were used having dimensions and activation function as one hundred and five and ‘relu’, fifty-two and ‘relu’, twenty-six and ‘relu’, thirteen and ‘relu’, six and ‘relu’, three and ‘relu’, and lastly, two and ‘sigmoid’. The model is compiled using the loss function as ‘categorical\_crossentropy’, optimiser as adam, and metrics as accuracy. After training the neural network model with a max epoch count of one hundred, the prediction was made, and the result was converted into an array back from One-Hot-Encoded format. The phenotypic data is processed using SVM. Since the purpose was to use the result obtained in ensembling, the SVM was optimised using Grid Search to predict the best suitable parameters. The regularisation parameter was set to one thousand, gamma as 0.05, and kernel as ‘rbf’. The model was further trained using the scaled data from unzipping to obtain the prediction result. The flow chart in Figure \ref{17} depicts the various methodologies used to process the result. The phenotypic data of these individuals was processed to be used in conjunction with fMRI to ensemble the classification of ADHD. The fMRI data was processed using the Neural Network, and the phenotypic are processed using SVM. They were ensembled to yield a diagnosis that predicts ADHD based on an individual’s fMRI and phenotypic data. Varying the weightage depicts the strength of each parameter. For the testing, both parameters with equal weightage were used. Since the prediction class is binary, the results obtained from both the fMRI and phenotypic classification were multiplied with their weightage and added together. A threshold of 0.5 determined the prediction class. If the sum is less than or equal to the threshold, it belongs to class 0. If the sum is greater than this threshold, it belongs to class 1. The result obtained was further used to display the confusion matrix and other relevant visualisations.

\section{Validation of Precise Framework and Result Analysis}

The phenotypic dataset was classified using multiple machine learning algorithms. The Logistic Regression gave an accuracy of 86.13\%, Random Forest gave 84.15\%, Ada-Boost gave 69.30\%, Support Vector Machine gave 90.09\%, K-Nearest Neighbour gave 77.27\%, and Artificial Neural Network gave 85.14\%. Figure \ref{18} depicts the detailed analysis. Diagnosing ADHD based on the dataset of 39 individuals from nilearn showed an accuracy of 75\% for Artificial Neural Network on fMRI, 100\% for SVM on phenotypic, and 100\% when combined using both of these classifiers. Figure \ref{19} depicts the detailed analysis. Diagnosing ADHD based on the dataset of 400 individuals from NITRC gave an accuracy of 86.67\% for Artificial Neural Network on fMRI, 99.16\% for SVM on phenotypic and 92.5\% when combined using both of these classifiers. Figure \ref{20} depicts the detailed analysis.

People have tried to develop multiple binary and multiclass classifiers based on either rs-fMRI or phenotypic data of an individual or both to diagnose and classify ADHD. After studying the current technological developments, we experimented with multiple techniques to further advance the diagnosis and found many exciting observations and critical areas of improvement for research development. Therefore, we started working on classifying ADHD using the phenotypic data of individuals provided by NITRC under the ADHD-200 competition. As a result, we achieved an accuracy of 90.09\% using Support vector machines for unified data and could be increased further if split into sections. However, upon detailed analysis of this data, we found a massive variation in classification based on the source of the data and the methodology used to calculate various parameters. Moreover, it convinced us that the data was not enough to apply methods for medical application and clinical purposes.\\

Therefore, we looked into MRI, hoping it could provide us with a consistent evaluation platform for accurately diagnosing ADHD. Looking at the rs-fMRI data supplied by the nilearn library, we started modifying current work with a better neural network to diagnose ADHD based on these rs-fMRI and successfully achieved an accuracy of 75\%. The next task was to integrate the diagnosis using rs-fMRI with the classification using phenotypic to provide the best result for the application in medical and clinical areas. We achieve an astonishing accuracy of 100\% accuracy with SVM and accuracy of 100\% when ensembled with equal weightage due to accurate prediction of non-ADHD by the neural network. However, the nilearn library consisted of only 40 individuals at the time at version nilearn-0.7.1 and hence still needed to be proven for larger applications. Therefore, we hosted the rs-fMRI images and phenotypic privately on the cloud, taken from NITRC, and wrote a script to automate the process to work with a larger count of individuals. We managed to host 824 individuals, and upon removing the discrepancies, were reduced to 400. The neural network on fMRI produced an accuracy of 86.67\%, and the SVM on the phenotypic yielded an accuracy of 99.16\%, both contributing to the ensemble in equal weightage with an accuracy of 92.5\%. Figure \ref{21} demonstrates the comparison of accuracy achieved using all of the classifications models as mentioned earlier [38].
Figure \ref{21} description: The classifiers in blue are based on the phenotypic dataset of 505 individuals. These six classifiers represent Logistic Regression, Random Forest, Adaptive-Boosting, Support Vector Machine, K-Nearest Neighbour, and Artificial Neural Network. The classifiers in red are based on the dataset of 39 individuals obtained from the nilearn library. The classifiers in yellow are based on the dataset of 400 individuals obtained from the NITRC dataset. For both these colours, ANN-MRI represents the Artificial Neural Network on fMRI data, SVM represents the Support Vector Machine on phenotypic data, and the Ensemble represents classification used in both the data.

\begin{table*}[htbp]
	\caption{Comparison of iPAL with Existing Works}
	\label{table_example}
	\centering
	\begin{tabular}{llllllll}
		\hline
		\textbf{Works}&\textbf{Dataset}&Subjects&Features&Classifier& \textbf{Accy (\%)}&\textbf{Prcs (\%)}&\textbf{F-1}\\
		\hline		 \hline
		Khare et.al.[39]&EEG&61&VHERS&ELM-SIG&99&100&99\\
		Tenev et.al.[40]&EEG&67&FFT&SVM &82.3&-&-\\
		Moghaddari et.al.[41]&EEG&31&Rhythms \& CNN&CNN &98.48&-&98.49\\
		Snyder et.al.[42]&EEG&97&Mean Power&- &89&87&-\\
		Karimu et.al.[43]&CWT&20&Scalogram&MEFM&99&98&-\\
		Boroujeni et.al.[44]&EEG&50&FD,CD,EN&SVM&96&98&-\\
		Proposed iPAL&phenotypic &400&meanFD,DVARS&Ensemble&100&99&99\\
		& \& fMRI&400&rmsFD&&&&\\
		\hline
	\end{tabular}
\end{table*}

\section{Conclusion and Future Direction of Current Research}
Multiple regions data is collected and factors have been considered for improvement during the analysis of research framework before a computational diagnosis of ADHD. This will be beneficial for real-world medical and clinical applications. Huge variations exist between the data provided by numerous websites and the measures adopted to compute various parameters. The prediction using phenotypic is precise comparatively. The proposed research paradigm confirms that working with fMRI can provide a crucial relationship with ADHD if the data is provided with minimalist noise and follows unified techniques for measuring parameters. Meanwhile, the phenotypic does a much better job at diagnosing ADHD. A large data can provide us meaningful insights and represent the relation between various measures and given sufficient data. Hence a suitable standard needs to be adopted for better diagnosis.\\

Based on the present conclusive research, it is obvious that the precise diagnosis is required for mitigating the impact of attention deficit/hyperactivity disorder. Hence, it is required to design or form the paradigm in which, automatic diagnosis is possible with current approaches for precise detection point of view. The diagnosis would be possible for patients of remote locations. An intelligent framework is in the future plan, which functions the all processes such as data collection from patients, automatic diagnosis and providing treatment synchronously to the individuals with time optimization. There will be an advantage is that the patient will be able to take better and reliable treatment without failure. This type of medical care paradigm will help the patient to provide the treatment at low cost. The people would be conscious for similar kind of disorders and would be smarter in self diagnosis after getting the frequent solutions.

% Can use something like this to put references on a page
% by themselves when using endfloat and the captionsoff option.
%\ifCLASSOPTIONcaptionsoff
  
%\fi

%\bibliographystyle{IEEEtran}
%\bibliography{ref}
\section*{References}
[1] J. Berrezueta-Guzman, V. E. Robles-Bykbaev, I. Pau, F. Pesántez-Avilés and M. -L. Martín-Ruiz, "Robotic Technologies in ADHD Care: Literature Review," in IEEE Access, vol. 10, pp. 608-625, 2022. doi: 10.1109/ACCESS.2021.3137082.

[2] Y. Gu, S. Miao, J. Yang and X. Li, "ADHD Children Identification With Multiview Feature Fusion of fNIRS Signals," in IEEE Sensors Journal, vol. 22, no. 13, pp. 13536-13543, 1 July1, 2022. doi: 10.1109/JSEN.2022.3168488.

[3] S. Pei, C. Wang, S. Cao and Z. Lv, "Data Augmentation for fMRI-Based Functional Connectivity and its Application to Cross-Site ADHD Classification," in IEEE Transactions on Instrumentation and Measurement, vol. 72, pp. 1-15, 2023, Art no. 2501015. doi: 10.1109/TIM.2022.3232670.

[4] Ankrah, Elizabeth A., et al. "Me, My Health, and My Watch: How Children with ADHD Understand Smartwatch Health Data." ACM Transactions on Computer-Human Interaction (2022).

[5] Tavakoulnia, Arya, et al. "Designing a wearable technology application for enhancing executive functioning skills in children with ADHD." Adjunct Proceedings of the 2019 ACM International Joint Conference on Pervasive and Ubiquitous Computing and Proceedings of the 2019 ACM International Symposium on Wearable Computers. 2019.

[6]“Epidemiology”,https://adhd-institute.com/burden-of-adhd/epidemiology/, march 2021.

[7] R. Bansal, L. H. Staib, D. Xu, H. Zhu, and B. S. Peterson, “Statistical analyses of brain surfaces using Gaussian random fields on 2-D manifolds,” IEEE Transactions on Medical Imaging, vol. 26, no. 1, pp. 46–57, 2007.

[8] L. Shao, Y. You, H. Du, and D. Fu, “Classification of ADHD with fMRI data and multi-objective optimization,” Computer Methods and Programs in Biomedicine, vol. 196, p. 105676, 2020.

[9] A. Kushwaha, “Attention deficit hyperactivity disorder prediction using machine learning,” International Journal of Advanced Trends in Computer Science and Engineering, vol. 9, pp. 5296–5302, 08 2020.

[10] P. Bellec, C. Chu, F. Chouinard-Decorte, Y. Benhajali, D. S. Margulies, and R. C. Craddock, “The neuro bureau ADHD-200 preprocessed repository,” Neuroimage, vol. 144, pp. 275–286, 2017.

[11] Mennes et al,“The ADHD-200 sample,” https://fcon\textunderscore1000.projects.nitrc.org/indi/adhd200/.

[12] H. RaviPrakash, A. Watane, S. Jambawalikar, and U. Bagci, “Deep learning for functional brain connectivity: Are we there yet?” Deep Learning and Convolutional Neural Networks for Medical Imaging and Clinical Informatics, pp. 347–365, 2019.

[13] A. Riaz, E. Alonso, and G. Slabaugh, “Phenotypic integrated framework for classification of ADHD using fMRI,” in International Conference on Image Analysis and Recognition. Springer, 2016, pp. 217–225.

[14] M. N. I. Qureshi, B. Min, H. J. Jo, and B. Lee, “Multiclass classification for the differential diagnosis on the ADHD subtypes using recursive feature elimination and hierarchical extreme learning machine: structural MRI study,” PloS one, vol. 11, no. 8, p. e0160697, 2016. pp:- 22-25.

[15] N. Sethu and R. Vyas, “Overview of machine learning methods in ADHD prediction,” in Advances in Bioengineering. Springer, 2020, pp. 51–71.

[16] M. Chen, H. Li, J. Wang, J. R. Dillman, N. A. Parikh, and L. He, “A multichannel deep neural network model analyzing multiscale functional brain connectome data for attention deficit hyperactivity disorder detection,” Radiology: Artificial Intelligence, vol. 2, no. 1, 2019.

[17] V. Pereira-Sanchez and F. X. Castellanos, “Neuroimaging in attention-deficit/hyperactivity disorder,” Current opinion in psychiatry, vol. 34, no. 2, p. 105, 2021.

[18] I. Elujide, S. G. Fashoto, B. Fashoto, E. Mbunge, S. O. Folorunso, and J. O. Olamijuwon, “Application of deep and machine learning techniques for multi-label classification performance on psychotic disorder diseases,” Informatics in Medicine Unlocked, vol. 23, p. 100545, 2021.

[19] S. Itani, F. Lecron, and P. Fortemps, “A multi-level classification framework for multi-site medical data: Application to the ADHD-200 collection,” Expert Systems with Applications, vol. 91, pp. 36–45, 2018.

[20] G. Tostaeva, “Using neural networks for a functional connectivity classification of fMRI data,” https://towardsdatascience.com/using-neural-networks-for-a-functional-connectivity-classification-of-fmridata-ff0999057bc6, Dec 2019.

[21] A. Abraham, F. Pedregosa, M. Eickenberg, P. Gervais, A. Mueller, J. Kossaifi, A. Gramfort, B. Thirion, and G. Varoquaux, “Machine learning for neuroimaging with scikit-learn,” Frontiers in neuroinformatics, vol. 8, p. 14, 2014.

[22] Y. Du, Z. Fu, and V. D. Calhoun, “Classification and prediction of brain disorders using functional connectivity: promising but challenging,” Frontiers in neuroscience, vol. 12, p. 525, 2018.

[23] A. M. Mowinckel, D. Alnæs, M. L. Pedersen, S. Ziegler, M. Fredriksen, T. Kaufmann, E. Sonuga-Barke, T. Endestad, L. T. Westlye, and G. Biele, “Increased default-mode variability is related to reduced taskperformance and is evident in adults with ADHD,” NeuroImage: Clinical, vol. 16, pp. 369–382, 2017.

[24] F. Pedregosa, G. Varoquaux, A. Gramfort, V. Michel, B. Thirion, O. Grisel, M. Blondel, P. Prettenhofer, R.Weiss, V. Dubourg et al., “Scikit-learn: Machine learning in python,” the Journal of machine Learning research, vol. 12, pp. 2825–2830, 2011.

[25] D. Tomasi and N. D. Volkow, “Abnormal functional connectivity in children with attention-deficit/hyperactivity disorder,” Biological psychiatry, vol. 71, no. 5, pp. 443–450, 2012.

[26] G. Varoquaux, S. Sadaghiani, P. Pinel, A. Kleinschmidt, J.-B. Poline, and B. Thirion, “A group model for stable multi-subject ICA on fMRI datasets,” Neuroimage, vol. 51, no. 1, pp. 288–299, 2010.

[27] H. Yang, Q.-Z. Wu, L.-T. Guo, Q.-Q. Li, X.-Y. Long, X.-Q. Huang, R. C. Chan, and Q.-Y. Gong, “Abnormal spontaneous brain activity in medication-naive ADHD children: a resting state fMRI study,” Neuroscience letters, vol. 502, no. 2, pp. 89–93, 2011.

[28] Abraham, A., Pedregosa, F., Eickenberg, M., Gervais, P., Mueller, A., Kossaifi, J., Gramfort, A., Thirion, B., \& Varoquaux, G. (2014). Machine learning for neuroimaging with scikit-learn. Frontiers in Neuroinformatics, 8. https://doi.org/10.3389/fninf.2014.00014.

[29] L. E. Johnson and J. M. Conrad, “A survey of technologies utilized in the treatment and diagnosis of attention deficit hyperactivity disorder,” in 9th IEEE Annual Ubiquitous Computing, Electronics \& Mobile Communication Conference (UEMCON). IEEE, 2018, pp. 819–824.

[30] B. Sorger and R. Goebel, “Chapter 21 - real-time fMRI for brain-computer interfacing,” in Brain-Computer Interfaces, ser. Handbook of Clinical Neurology, N. F. Ramsey and J. del R. Millán, Eds. Elsevier, 2020, vol. 168, pp. 289–302.

[31] S. A. Huettel, “Functional MRI (fMRI),” in Encyclopedia of Spectroscopy and Spectrometry,second edition ed., J. C. Lindon, Ed. Oxford: Academic Press, 2010, pp. 741–748.

[32] E. B. Johnson and S. Gregory, “Chapter 6-huntington’s disease: Brain imaging in huntington’s disease,” in Brain Imaging, ser. Progress in Molecular Biology and Translational Science, 2019, vol. 165, pp. 321–369.

[33] A. Zalesky, A. Fornito, and E. Bullmore, “On the use of correlation as a measure of network connectivity,” NeuroImage, vol. 60, no. 4, pp. 2096–2106, 2012.

[34] F. Pedregosa, G. Varoquaux, A. Gramfort, V. Michel, B. Thirion, O. Grisel, M. Blondel, P. Prettenhofer, R.Weiss, V. Dubourg, J. Vanderplas, A. Passos, D. Cournapeau, M. Brucher, M. Perrot, and E. Duchesnay, “Scikit-learn: Machine learning in Python,” Journal of Machine Learning Research, vol. 12, pp. 2825–2830, 2011.

[35] C. McNorgan, C. Judson, D. Handzlik, and J. G. Holden, “Linking ADHD and behavioral assessment through identification of shared diagnostic task-based functional connections,” Frontiers in Physiology, vol. 11, p. 1595, 2020.

[36] H. Zhang, Y. Zhao,W. Cao, D. Cui, Q. Jiao, W. Lu, H. Li, and J. Qiu, “Aberrant functional connectivity in resting state networks of ADHD patients revealed by independent component analysis,” BMC neuroscience, vol. 21, no. 1, pp. 1–11, 2020.

[37] S. Itani, M. Rossignol, F. Lecron, and P. Fortemps, “Towards interpretable machine learning models for diagnosis aid: a case study on attention deficit/hyperactivity disorder,” PloS one, vol. 14, no. 4, p. e0215720, 2019.

[38] Sharma, Yogesh, and Bikesh Kumar Singh. "Attention deficit hyperactivity disorder detection in children using multivariate empirical EEG decomposition approaches: A comprehensive analytical study." Expert Systems with Applications 213 (2023): 119219.

[39] S. K. Khare, N. B. Gaikwad and V. Bajaj, "VHERS: A Novel Variational Mode Decomposition and Hilbert Transform-Based EEG Rhythm Separation for Automatic ADHD Detection," in IEEE Transactions on Instrumentation and Measurement, vol. 71, pp. 1-10, 2022, Art no. 4008310. doi: 10.1109/TIM.2022.3204076.

[40] Tenev, Aleksandar, et al. "Machine learning approach for classification of ADHD adults." International Journal of Psychophysiology 93.1 (2014): 162-166.

[41] Moghaddari, Majid, Mina Zolfy Lighvan, and Sebelan Danishvar. "Diagnose ADHD disorder in children using convolutional neural network based on continuous mental task EEG." Computer Methods and Programs in Biomedicine 197 (2020): 105738.

[42] Snyder, Steven M., et al. "Integration of an EEG biomarker with a clinician's ADHD evaluation." Brain and behavior 5.4 (2015): e00330.

[43] Yaghoobi Karimu, Reza, and Sassan Azadi. "Diagnosing the ADHD using a mixture of expert fuzzy models." International Journal of Fuzzy Systems 20.4 (2018): 1282-1296.

[44] Boroujeni, Yasaman Kiani, Ali Asghar Rastegari, and Hamed Khodadadi. "Diagnosis of attention deficit hyperactivity disorder using non‐linear analysis of the EEG signal." IET systems biology 13.5 (2019): 260-266.
\newpage
\section*{About the Authors}
\begin{minipage}[htbp]{\columnwidth}
	\begin{wrapfigure}{l}{1.0in}
		\vspace{-0.4cm}
		\includegraphics[width=1.0in,keepaspectratio]{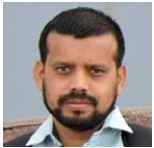}
		\vspace{-0.3cm}
	\end{wrapfigure}
	\noindent
	\textbf{Abhishek Sharma} received his B.E. in Electronics Engineering from Jiwaji University, Gwalior, India, and Ph.D. in Embedded Systems from the University of Genoa, Italy. He is presently working as an Assistant Professor in the Department of ECE at LNM Institute of Information Technology, Jaipur, India. He is also a member of IEEE, Computer Society and Consumer Electronics Society, a Lifetime Member of the Indian Society for Technical Education, a Lifetime member of Advanced Computing Society, India. In the present institute, he is the coordinator of the ARM university partner program. He is also the center leader of the LNM Smart Technology Center (L-CST). His research interests include real-time systems and embedded systems.
	\end{minipage}
	
	\vspace{1.8cm}

    \begin{minipage}[htbp]{\columnwidth}
	\begin{wrapfigure}{l}{1.0in}
		\vspace{-0.4cm}
		\includegraphics[width=1.0in,keepaspectratio]{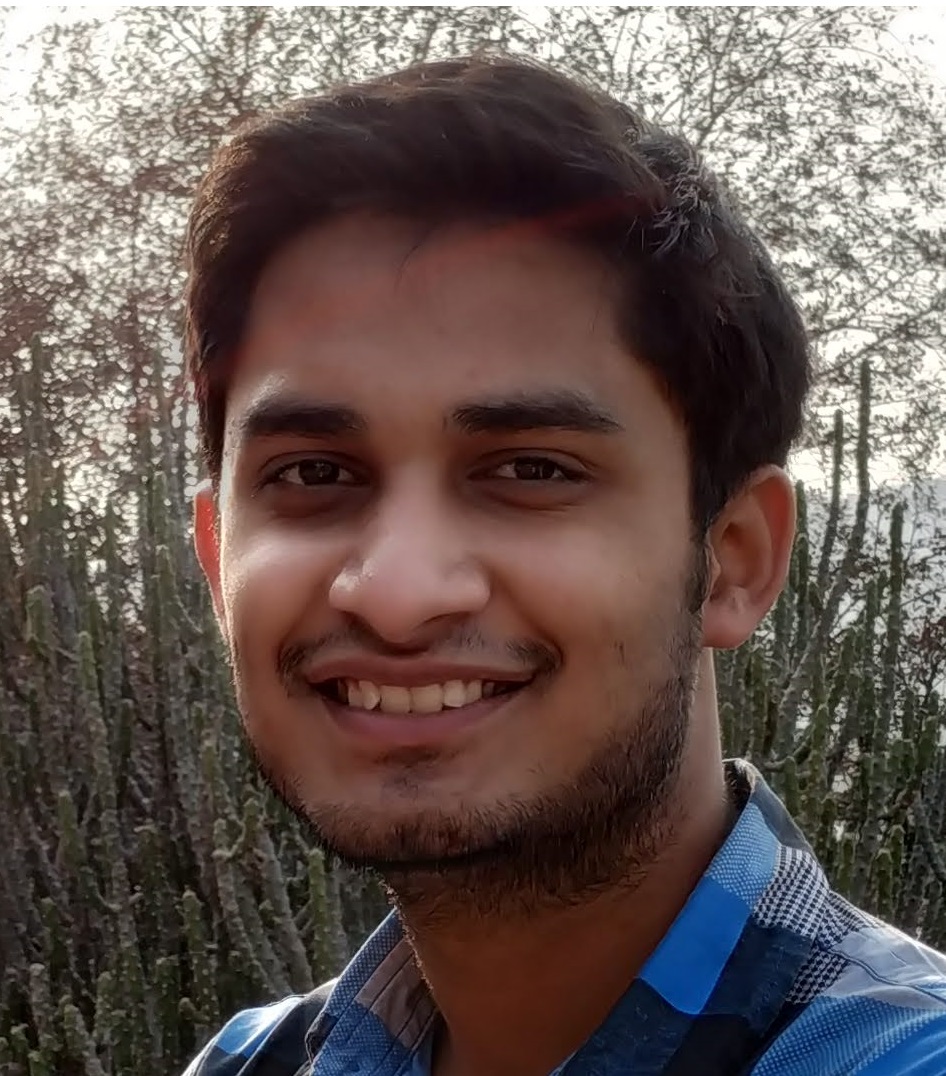}
		\vspace{-0.3cm}
	\end{wrapfigure}
	\noindent
	\textbf{Arpit Jain} is currently a pre-final year student at The LNM Institute of Information Technology and is pursuing a B.Tech in Computer Science and Engineering. He is currently a mentor and earlier coordinator for MWEP(Sankalp Club), a senior member of Phoenix Club(Robotics), Astronomy Club, Google Developers Group (GDG Lucknow). He is currently working as an intern for VMware India Private Limited. This paper is his first significant contribution in the field of Neuroimaging and ADHD. His major area of study includes Machine Learning, Artificial Intelligence, Social Network Analysis, Database Management, Wireless Sensor Network, Network-on-Chip, Image Processing and Software Development Engineering.
	\end{minipage}
	
	\vspace{1.8cm}
	
		\begin{minipage}[htbp]{\columnwidth}
	\begin{wrapfigure}{l}{1.0in}
		\vspace{-0.4cm}
		\includegraphics[width=1.0in,keepaspectratio]{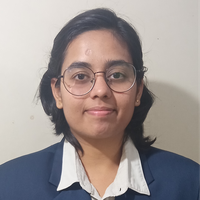}
		\vspace{-0.3cm}
	\end{wrapfigure}
	\noindent
	\textbf{Shubhangi Sharma} is currently a pre-final year student at the LNM Institute of Information Technology and is pursuing a B.Tech degree in  Electronics and Communication  Engineering. This paper is her first significant work in the area of machine learning and neuroimaging. Her areas of interest include Machine Learning, Artificial intelligence, Internet of things, embedded systems and software development.
	\end{minipage}
	
	\vspace{1.8cm}
		\begin{minipage}[htbp]{\columnwidth}
	\begin{wrapfigure}{l}{1.0in}
		\vspace{-0.4cm}
		\includegraphics[width=1.0in,keepaspectratio]{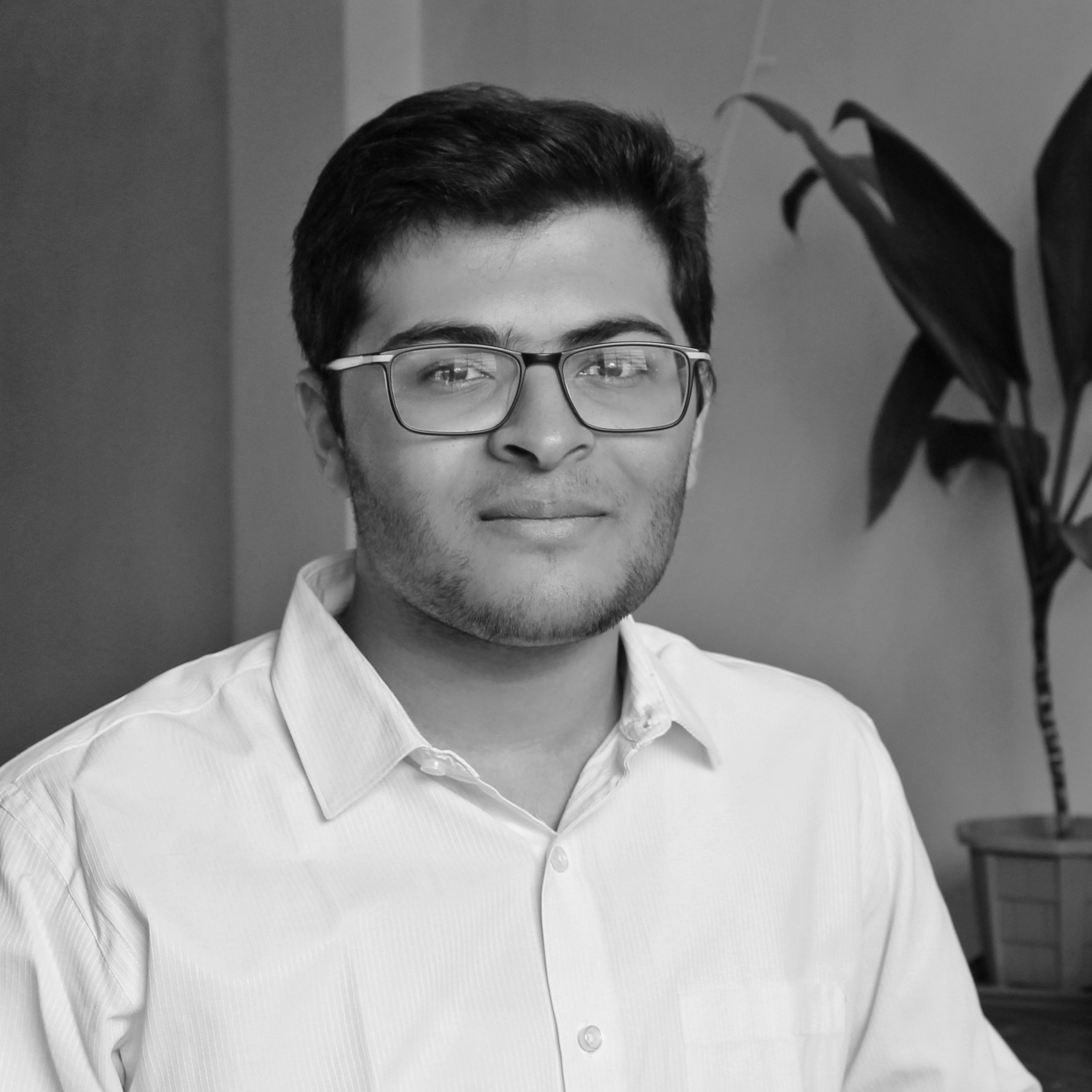}
		\vspace{-0.3cm}
	\end{wrapfigure}
	\noindent
	\textbf{Ashutosh Gupta} is currently a pre-final year undergraduate student at the LNM Institute of Information Technology, Jaipur, India, pursuing a degree in Computer Science Engineering. The work presented here is his first paper in the area of neuroimaging and data science. His areas of interest include Machine Learning, Software development, design strategies in machine learning and software development , and research in data science. 
	\end{minipage}
	
	\vspace{1.8cm}
\begin{minipage}[htbp]{\columnwidth}
	\begin{wrapfigure}{l}{1.0in}
		\vspace{-0.4cm}
		\includegraphics[width=1.1in,keepaspectratio]{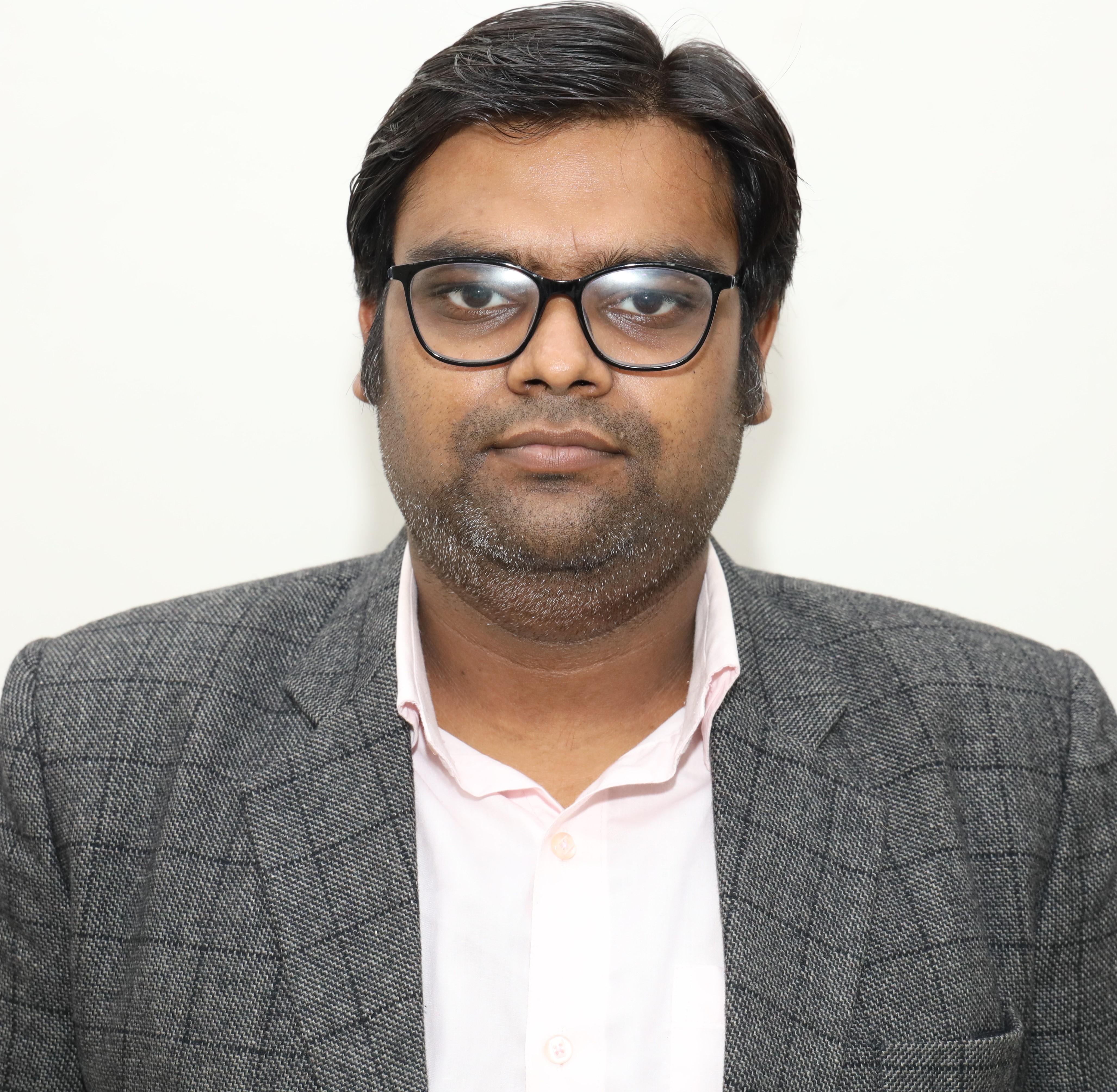}
		%\vspace{-0.3cm}
	\end{wrapfigure}
	\noindent
	\textbf{Prateek Jain} (M'18) earned his B.E. degree in Electronics Engineering from Jiwaji University, India in 2010 and Master degree from ITM University Gwalior. He obtained PhD from Malaviya National Institute of Technology, Jaipur. He was awarded from MHRD fellowship during 2016-2020. He was faculty member in school of electronics engg. (SENSE), VIT AP University. Currently, he is an Assistant Professor in Nirma University, Ahmedabad, India. His current research interest includes Real-time system design, Low power VLSI design, Biomedical Systems and Instrumentation. He is an author of 25 peer-reviewed publications. He is a regular reviewer of 12 journals and 10 conferences. He was resource person in reputed universities for technical programs.
	\end{minipage}
	\vspace{0.3cm}

\begin{minipage}[htbp]{\columnwidth}
	%\begin{IEEEbiography}
	%[{\includegraphics[height=1.25in,keepaspectratio]{Saraju_Mohanty}}] 
	\begin{wrapfigure}{l}{1.00in}
		\vspace{-0.3cm}
		\includegraphics[width=1.0in,keepaspectratio]{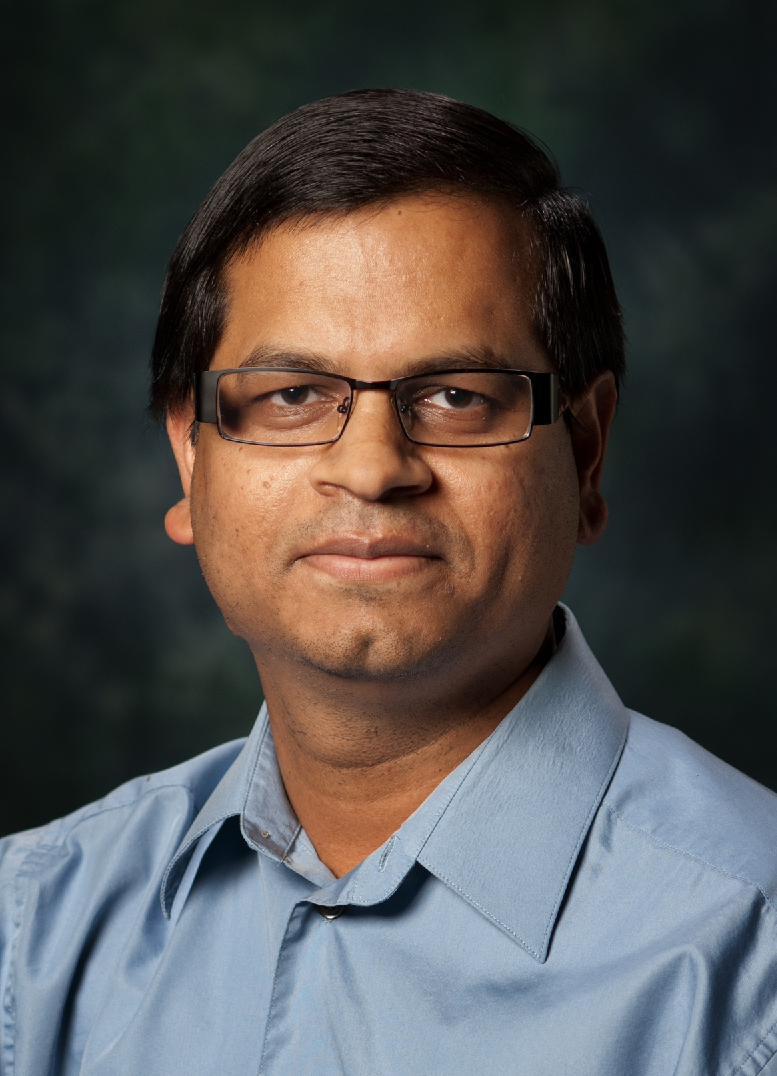}
		\vspace{-0.5cm}
	\end{wrapfigure}
	\noindent
	\textbf{Saraju P. Mohanty} (Senior Member, IEEE) received the bachelor’s degree (Honors) in electrical engineering from the Orissa University of Agriculture and Technology, Bhubaneswar, in 1995, the master’s degree in Systems Science and Automation from the Indian Institute of Science, Bengaluru, in 1999, and the Ph.D. degree in Computer Science and Engineering from the University of South Florida, Tampa, in 2003. He is a Professor with the University of North Texas. His research is in ``Smart Electronic Systems’’ which has been funded by National Science Foundations (NSF), Semiconductor Research Corporation (SRC), U.S. Air Force, IUSSTF, and Mission Innovation. He has authored 450 research articles, 5 books, and 9 granted and pending patents. His Google Scholar h-index is 48 and i10-index is 209 with 10,500 citations. He is regarded as a visionary researcher on Smart Cities technology in which his research deals with security and energy aware, and AI/ML-integrated smart components. He introduced the Secure Digital Camera (SDC) in 2004 with built-in security features designed using Hardware Assisted Security (HAS) or Security by Design (SbD) principle. He is widely credited as the designer for the first digital watermarking chip in 2004 and first the low-power digital watermarking chip in 2006. He is a recipient of 16 best paper awards, Fulbright Specialist Award in 2020, IEEE Consumer Electronics Society Outstanding Service Award in 2020, the IEEE-CS-TCVLSI Distinguished Leadership Award in 2018, and the PROSE Award for Best Textbook in Physical Sciences and Mathematics category in 2016. He has delivered 18 keynotes and served on 14 panels at various International Conferences. He has been serving on the editorial board of several peer-reviewed international transactions/journals, including IEEE Transactions on Big Data (TBD), IEEE Transactions on Computer-Aided Design of Integrated Circuits and Systems (TCAD), IEEE Transactions on Consumer Electronics (TCE), and ACM Journal on Emerging Technologies in Computing Systems (JETC). He has been the Editor-in-Chief (EiC) of the IEEE Consumer Electronics Magazine (MCE) during 2016-2021. He served as the Chair of Technical Committee on Very Large Scale Integration (TCVLSI), IEEE Computer Society (IEEE-CS) during 2014-2018 and on the Board of Governors of the IEEE Consumer Electronics Society during 2019-2021. He serves on the steering, organizing, and program committees of several international conferences. He is the steering committee chair/vice-chair for the IEEE International Symposium on Smart Electronic Systems (IEEE-iSES), the IEEE-CS Symposium on VLSI (ISVLSI), and the OITS International Conference on Information Technology (OCIT). He has mentored 3 post-doctoral researchers, and supervised 14 Ph.D. dissertations, 26 M.S. theses, and 18 undergraduate projects.
\end{minipage}

\end{document}